\documentclass[10pt,twocolumn,letterpaper]{article}

\usepackage{cvpr}              

\usepackage[accsupp]{axessibility}  

\definecolor{cvprblue}{rgb}{0.21,0.49,0.74}
\usepackage[pagebackref,breaklinks,colorlinks,allcolors=cvprblue]{hyperref}

\def\paperID{7949} 
\def\confName{CVPR}
\def\confYear{2025}

\usepackage{float}
\usepackage{graphicx}
\usepackage{multirow}
\usepackage{amsmath}
\usepackage{xcolor} 
\usepackage{caption}
\usepackage{algorithm}
\usepackage{algorithmic}
\usepackage{pifont} 
\usepackage{color, colortbl} 
\usepackage{arydshln} 
\usepackage{tabularx} 
\usepackage{footnote} 
\usepackage{booktabs} 
\newcommand{\customfootnotetext}[2]{{
\renewcommand{\thefootnote}{#1}
\footnotetext[0]{#2}}}
\newcommand{\vlsititle}{\includegraphics[width=0.15\textwidth, trim=0in 5in 2in 10in]{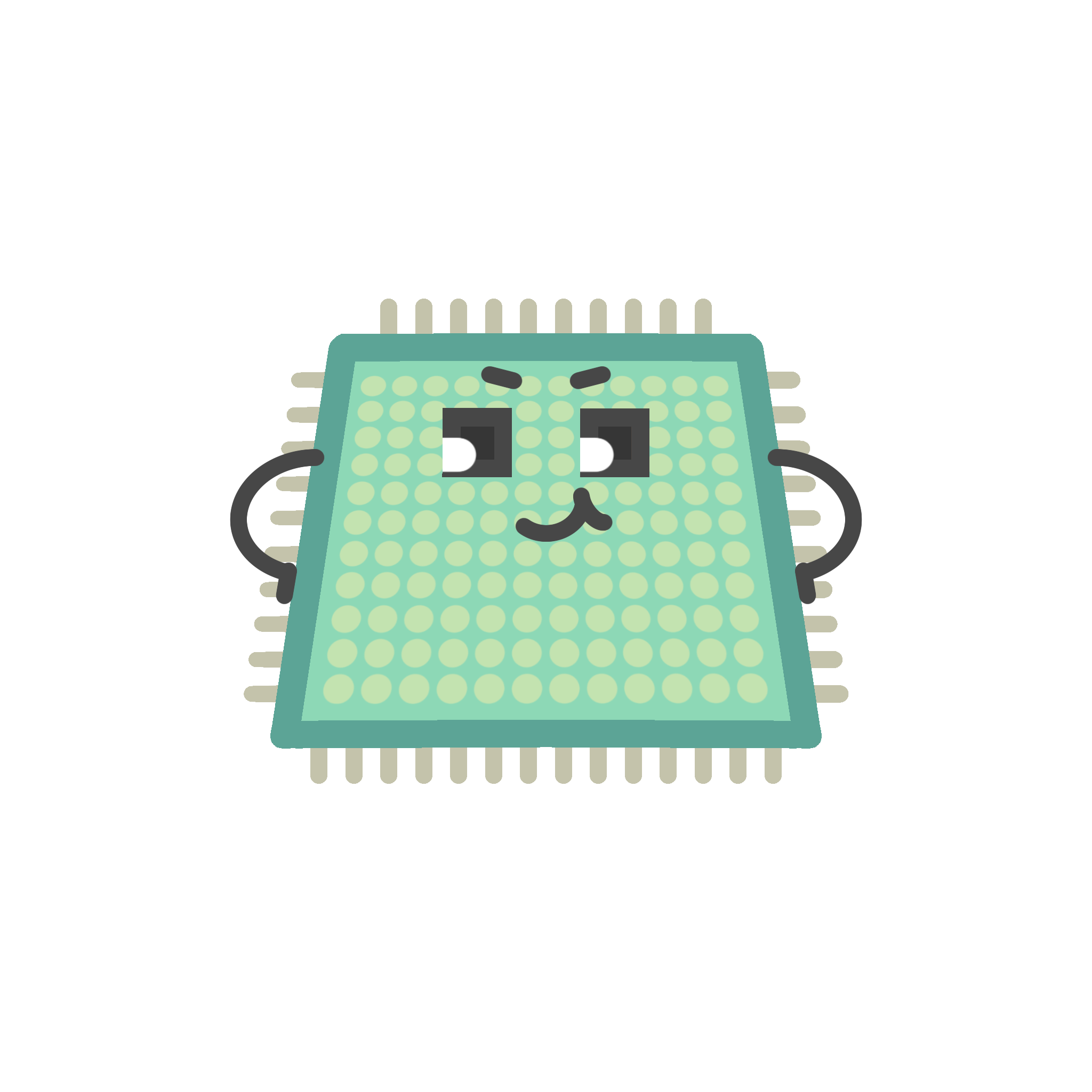}\hspace{0ex}} 
\newcommand{\vlsi}{\includegraphics[width=0.035\textwidth, trim=0in 5in 2in 10in]{figures/vlsi_emoji.png}\hspace{0.25ex}} 
\title{\vlsititle VLsI: Verbalized Layers-to-Interactions\\from Large to Small Vision Language Models}

\author{
Byung-Kwan Lee\textsuperscript{1,2$\dagger$}\quad\quad Ryo Hachiuma\textsuperscript{1}\quad\quad Yu-Chiang Frank Wang\textsuperscript{1,3}\\\\
Yong Man Ro\textsuperscript{2$\ddagger$}\quad\quad Yueh-Hua Wu\textsuperscript{1$\ddagger$}
\\\\
$\textsuperscript{\rm 1}$NVIDIA,\quad$\textsuperscript{\rm 2}$KAIST,\quad$\textsuperscript{\rm 3}$National Taiwan University
}

\begin{document}

\twocolumn[{%
\renewcommand\twocolumn[1][]{#1}%
\maketitle
\centering
\vspace{-7mm}
\includegraphics[width=0.9\textwidth]{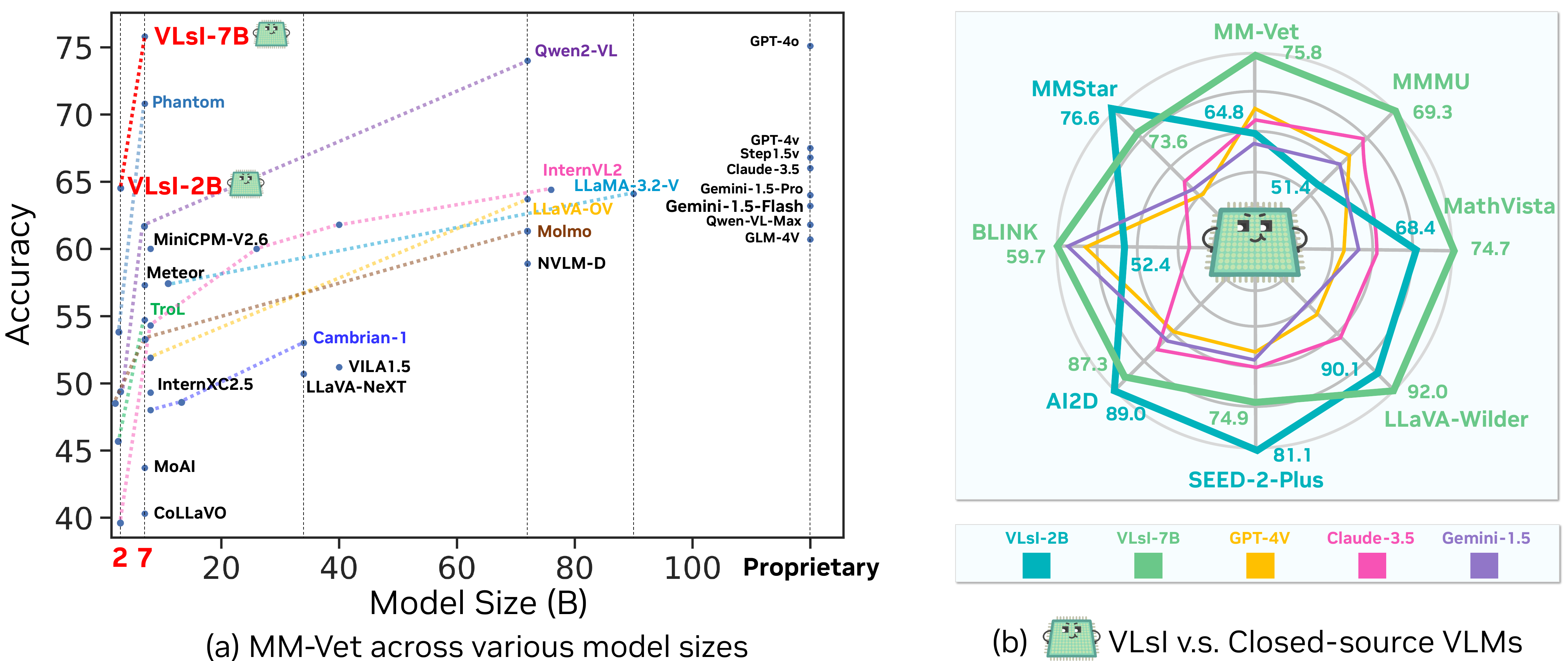}
\captionof{figure}{Performance overview of \vlsi VLsI on vision-language benchmarks. (a) Accuracy on MM-Vet~\cite{yu2023mm} for various model sizes, showing that \vlsi VLsI (2B and 7B) achieves competitive performance compared to proprietary closed-source VLMs. (b) Comparative evaluation on multiple challenging benchmarks, where \vlsi VLsI (green and blue) outperforms leading closed-source VLMs, including GPT-4V~\cite{gpttechnical}, Claude-3.5-Sonnet~\cite{claude3series2024}, and Gemini-1.5-Pro~\cite{team2023gemini}, highlighting its efficiency and effectiveness across diverse tasks.
}
\label{fig:1}
\vspace{2mm}
}]
\customfootnotetext{}{$\dagger$ Work Done during Internship.}
\customfootnotetext{}{$\ddagger$ Corresponding Author.}
\begin{abstract}
The recent surge in high-quality visual instruction tuning samples from closed-source vision-language models (VLMs) such as GPT-4V has accelerated the release of open-source VLMs across various model sizes. However, scaling VLMs to improve performance using larger models brings significant computational challenges, especially for deployment on resource-constrained devices like mobile platforms and robots. To address this, we propose \vlsi\textbf{VLsI}: \textbf{V}erbalized \textbf{L}ayer\textbf{s}-to-\textbf{I}nteractions, a new VLM family in 2B and 7B model sizes, which prioritizes efficiency without compromising accuracy. \vlsi VLsI leverages a unique, layer-wise distillation process, introducing intermediate ``verbalizers'' that map features from each layer to natural language space, allowing smaller VLMs to flexibly align with the reasoning processes of larger VLMs. This approach mitigates the training instability often encountered in output imitation and goes beyond typical final-layer tuning by aligning the small VLMs’ layer-wise progression with that of the large ones. We validate \vlsi VLsI across ten challenging vision-language benchmarks, achieving notable performance gains (11.0\% for 2B and 17.4\% for 7B) over GPT-4V without the need for model scaling, merging, or architectural changes. \href{https://byungkwanlee.github.io/VLsI-page}{Project Page}.
\end{abstract}    
\vspace{-5mm}
\section{Introduction}
\label{sec:intro}

The integration of large language models (LLMs) with vision-language models (VLMs) has significantly enhanced the interpretive and processing capabilities of visual systems~\cite{gptsyscard, cai2024internlm2, dubey2024llama, yang2024qwen2}. By leveraging architectures such as CLIP-aligned vision encoders~\cite{zhai2023sigmoid, chen2023internvl}, these VLMs have achieved unprecedented performance in understanding and responding to visual inputs. The core of these advancements is visual instruction tuning~\cite{liu2023visual, dai2023instructblip}, which pairs images with question-answer texts to provide VLMs with rich, instruction-based training. Closed-source VLMs like GPT-4V~\cite{gptsyscard} and Gemini-Pro~\cite{team2023gemini} have led the way, generating high-quality instruction tuning samples and setting new performance standards for visual language understanding.

In response, open-source VLMs of various sizes, including LLaVA-OneVision(OV)~\cite{li2024llava} and Qwen2-VL~\cite{wang2024qwen2vl}, have rapidly emerged. While these models demonstrate the advantage of scaling for performance gains in vision-language tasks, the computational cost of larger models presents a critical barrier to deployment in real-world, resource-limited settings, such as mobile devices and autonomous robots. Consequently, the challenge lies in designing high-performing, efficient VLMs capable of handling complex visual tasks without requiring extensive hardware resources.

Traditional approaches to address these constraints often involve adding specialized modules~\cite{lee2024moai} or modifying model architectures~\cite{lee2024phantom}. However, these methods introduce significant engineering complexity and can lead to compatibility issues during deployment, particularly for on-device applications where low-latency and memory efficiency are paramount. Furthermore, recent evaluations using benchmarks like MM-Vet~\cite{yu2023mm} and MMMU~\cite{yue2023mmmu} reveal that these structural modifications still struggle with advanced visual reasoning tasks. This raises the question: \textit{Can we achieve a similar or superior level of performance without scaling, merging, or architectural changes?}

To address this, we introduce \vlsi\textbf{VLsI}: \textbf{V}erbalized \textbf{L}ayer\textbf{s}-to-\textbf{I}nteractions, a new VLM family that leverages an innovative, natural language-based distillation process to efficiently transfer knowledge from large to small VLMs. Unlike traditional distillation methods, which often directly imitate outputs from a larger model, \vlsi VLsI introduces a layer-wise approach where each intermediate layer generates verbal responses in natural language space, enhancing interpretability and alignment with larger models. This is achieved through a three-step process: (1) the \textit{verbalization step}, which uses ``verbalizers'' to project intermediate features into the language space, making them interpretable as text-based responses; (2) the \textit{interaction step}, which performs adaptive layer matching to align the reasoning progression between large and small VLMs; and (3) the \textit{SFT step}, which finetunes the distilled VLMs for task-specific instruction-following responsiveness.

Our experiments validate \vlsi VLsI's effectiveness across ten challenging benchmarks, demonstrating significant performance gains of 11.0\% (2B model) and 17.4\% (7B model) over GPT-4V. Notably, these improvements are achieved without increasing model size, merging modules, or modifying the architecture. Our contributions are as:

\begin{figure*}[t!]
\vspace{-9mm}
    \centering
    {
    \begin{minipage}[t]{0.48\linewidth}
    \fbox{\includegraphics[width=\textwidth]{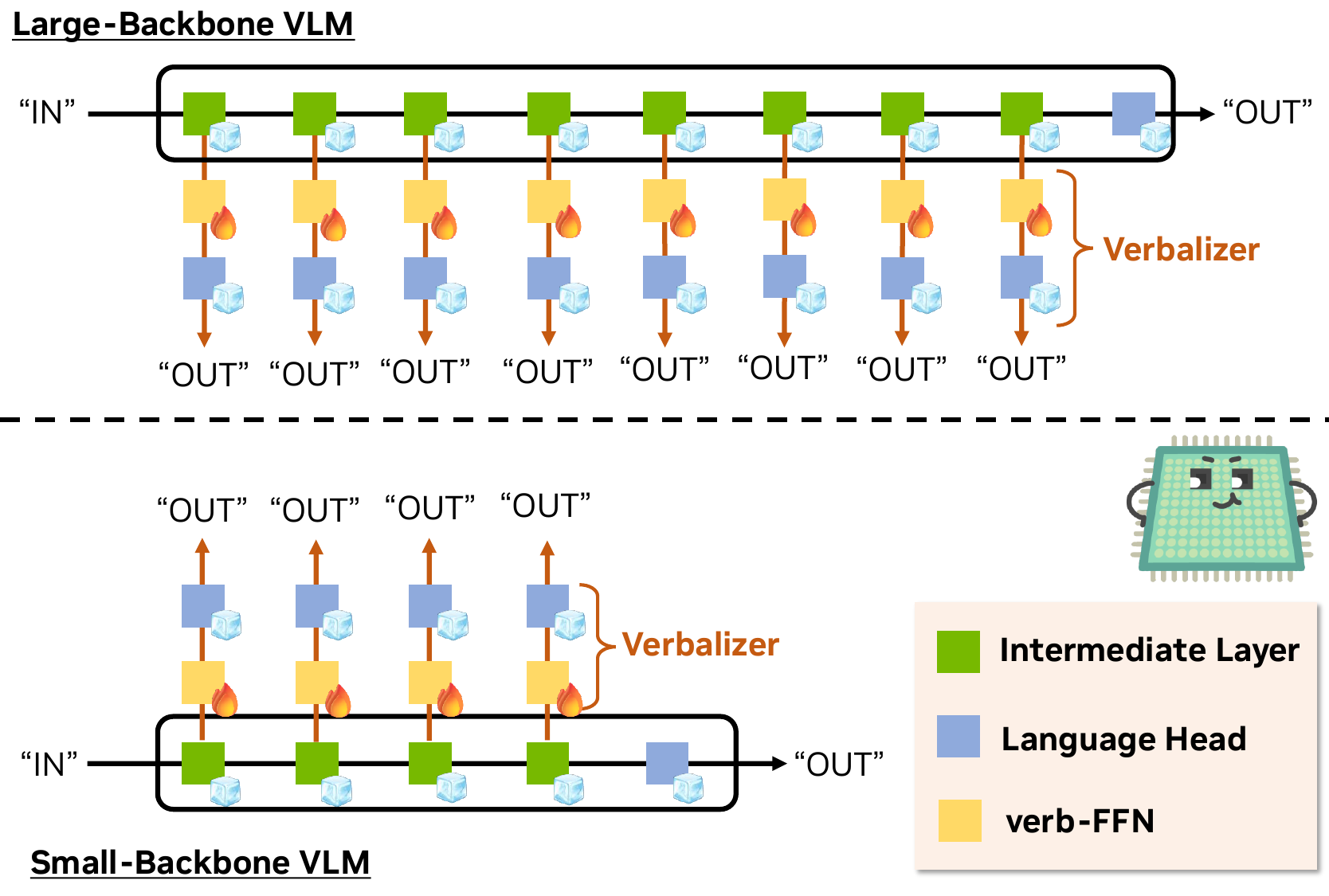}}
    \vspace*{-6mm}
    \caption*{\hspace{10mm}(a) \textit{Verbalization Step}}
    \end{minipage} \hspace{3mm}
    \begin{minipage}[t]{0.48\linewidth}
    \fbox{\includegraphics[width=\textwidth]{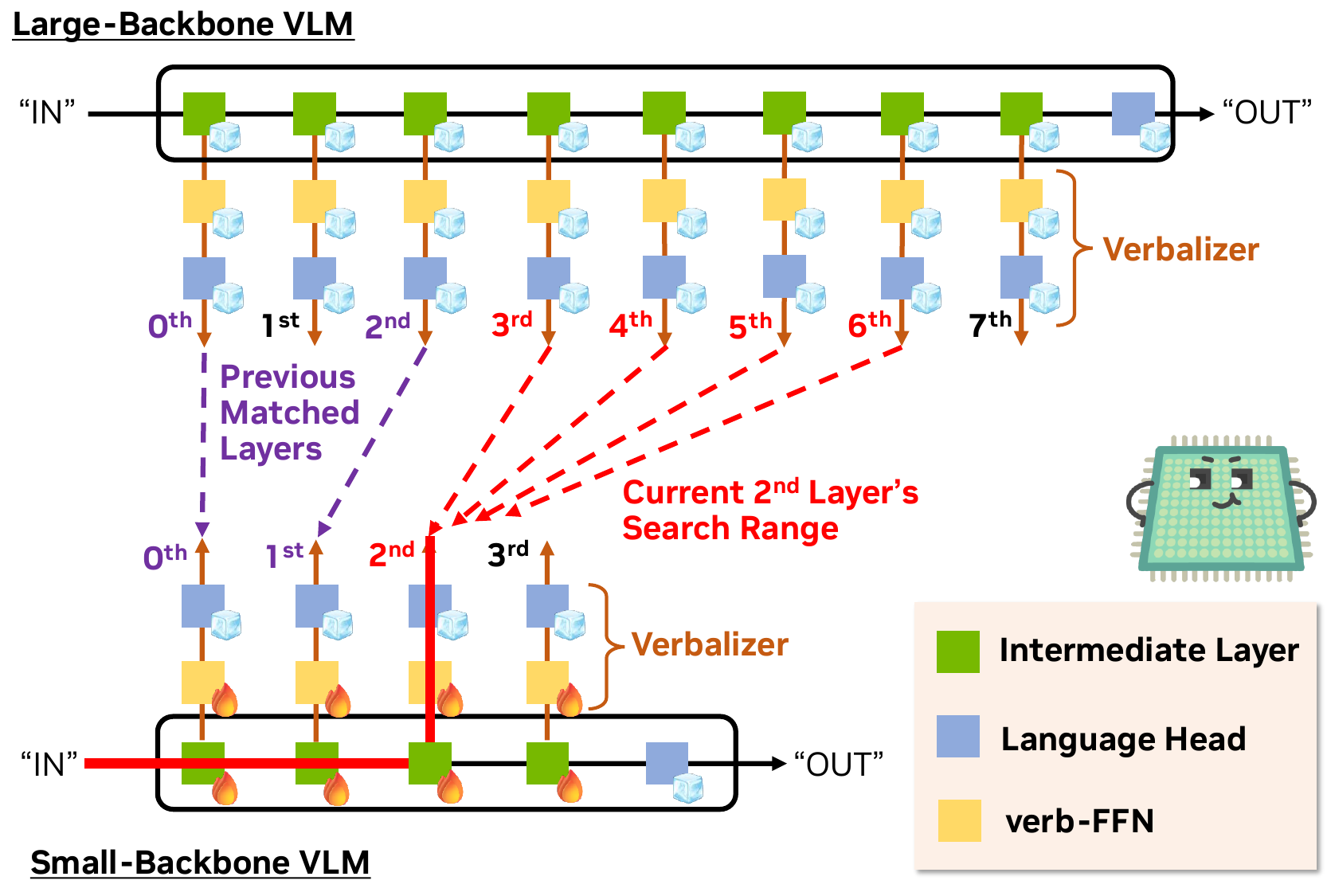}}
    \vspace*{-6mm}
    \caption*{(b) \textit{Interaction Step}}
    \end{minipage}
    }
    \vspace{-3mm}
    \caption{Illustration of the training process in \vlsi VLsI, showing (a) the verbalization step and (b) the interaction step. (a) In the verbalization step, intermediate layers in both the large- and small-backbone VLMs are equipped with a ``verbalizer'', allowing their outputs to be projected into natural language space. Autoregressive loss is applied to align these verbalized outputs with the target responses. (b) In the interaction step, each intermediate layer in the small-backbone VLM searches for a matching layer in the large backbone VLM within a specified range. For example, once the 2$^{\text{nd}}$ layer of the small VLM is matched with the 4$^{\text{th}}$ layer in the large VLM, the next matching search for the 3$^{\text{rd}}$ layer in the small VLM will proceed from the 5$^{\text{th}}$ to the 7$^{\text{th}}$ layers of the large VLM, ensuring progressive alignment. Note that, ``fireball'' icon represents parameters that are actively trained or updated, while the ``ice cube'' icon indicates untrained parameters.
}
    \label{fig:2}
    \vspace{-5mm}
\end{figure*}

\begin{itemize} 
    \item We introduce \vlsi VLsI, a new VLM family that applies natural language-based, layer-wise distillation, offering a scalable solution to high-performing yet efficient-scale VLMs without requiring scaling or structural changes. 
    \item We present sampling distribution-based dynamic layer matching before distillation, in order to address the layer number gap between large and small VLMs.
    \item \vlsi VLsI is easy to implement and adaptable across various model architectures, making it practical and deployable for real-world applications.
\end{itemize}

\section{Related Works}
\label{sec:related}

\paragraph{Evolution of Vision-Language Models.} The emergence of visual instruction tuning: LLaVA~\cite{liu2023visual} and  InstructBLIP~\cite{dai2023instructblip} initially brings in not only introducing slight variations of VLMs~\cite{laurenccon2023obelisc, chen2023shikra, bai2023qwen, zhu2023minigpt, li2023otter, 2023xtuner, zhang2023internlm}, but also curating high-quality visual instruction samples~\cite{chen2023sharegpt4v, chen2024allava, ye2023mplug, gao2023g, wang2024measuring, zhang2024beyond, fang2024vila}. Since that time, there has been a growing interest in enhancing visual understanding; thus the simple visual input technique of enlarging image resolution or dividing images into smaller sections with fixed or dynamic rules has got standardized~\cite{liu2024llavanext, mckinzie2024mm1, li2024mini, xu2024llava, chen2024far, ge2024convllava}. Furthermore, merging additional visual encoders~\cite{fang2023eva, oquab2023dinov2, kirillov2023segment, zhai2023sigmoid} or multiple computer vision models~\cite{cheng2022masked, minderer2023scaling, yang2022panoptic, du2021pp} into LLMs have also become a major focus~\cite{lu2024deepseek, goncharova2024omnifusion, zong2024mova, lee2024collavo, lee2024moai, shi2024eagle}. Besides, Meteor~\cite{lee2024meteor} employs an additional rational projector that embeds multifaceted reasoning information to cover diverse capabilities, including chart, diagram, document, and math reasoning. More recently, Cambrian-1~\cite{tong2024cambrian}, LLaVA-OneVision~\cite{li2024llava} InternVL2~\cite{chen2024far}, Molmo~\cite{deitke2024molmo}, and Qwen2-VL~\cite{wang2024qwen2vl} have released large scale models in order to follow or surpass the performances of closed-source VLMs. While these advancements are both rapid and impactful, relatively few studies focus on achieving high-performing yet efficient-scale VLMs within limited architectures. Furthermore, many current approaches rely heavily on GPT-based instruction datasets, and only the final layer learns the target responses in visual instruction tuning. In contrast, \vlsi VLsI presents how to effectively harness the capabilities of larger VLMs—already outperforming the closed-source ones—to transfer internally embedded knowledge from large to small VLMs.

\paragraph{Efficient Modeling Strategy.} In the field of lightweight LLMs, MobiLlama~\cite{thawakar2024mobillama}, OpenELM~\cite{mehta2024openelm}, MobileLLM~\cite{liu2024mobilellm} have leveraged various engineering techniques such as shared feed-forward network (FFN) design, layer-wise scaling, and embedding-language head sharing to efficiently reduce model parameters. Their primary objective is not to close the gap with closed-source LLMs but rather to reduce parameters impacting less performance degradation. In the field of VLMs, there are variations about how to utilize pretrained lightweight LLMs~\cite{zhang2024tinyllama, qwen, gu2023mamba} in order to make efficient-scale VLMs~\cite{zhou2024tinyllava, chu2023mobilevlm, chu2024mobilevlm, lin2024moe, qiao2024vl, zhao2024cobra}, but these works are also not a fundamental solution to embed more vision-language knowledge within limited structures. Notably, two VLM approaches, TroL~\cite{lee2024trol} and Phantom~\cite{lee2024phantom}, aim to expand learning own capabilities within limited structures by doubling forward propagation steps and enlarging the latent dimension without physically increasing model sizes, thereby showing large improvements. Unfortunately, these approaches face limitations such as key-value cache storage constraints and extensive architectural modifications, which may hinder direct application to real-world scenarios. Besides, a few works in distilling language models~\cite{liang2023less, muralidharan2024compact, sreenivas2024llm} have emerged, and TED~\cite{liang2023less} utilizes language head but it is not a version of LLM covering general tasks, and it does not handle layer number gap between large and small models. The other studies simply use the final layer distillation. For VLMs, LLaVA-MoD~\cite{shu2024llava} and LLaVA-KD~\cite{cai2024llava} have also been recently proposed, but they use the same way. Align-KD~\cite{feng2024align} distills both vision features and logits from the language head, while MoVE-KD~\cite{cao2025move} employs multiple vision encoders for large VLMs and a mixture of LoRA~\cite{hu2021lora} for small VLMs. Regarding the distillation metric, DistiLLM~\cite{ko2024distillm} and MiniLLM~\cite{gu2023minillm} emphasize the effectiveness of reverse KL divergence, whereas LLAVADI~\cite{xu2024llavadi} and Minitron~\cite{muralidharan2024compact} demonstrate that normal KL divergence is more effective. On the other hand, \vlsi VLsI makes layer-wise distillation process where we leverage natural language in order to make small VLMs mimic the reasoning progression of large VLMs across layers. We hope that incorporating natural language will facilitate smoother communication between large and small VLMs, alleviating the complexities of feature alignment.

\begin{figure*}[t!]
    \centering
    \vspace{-9mm}
    \includegraphics[width=\textwidth]{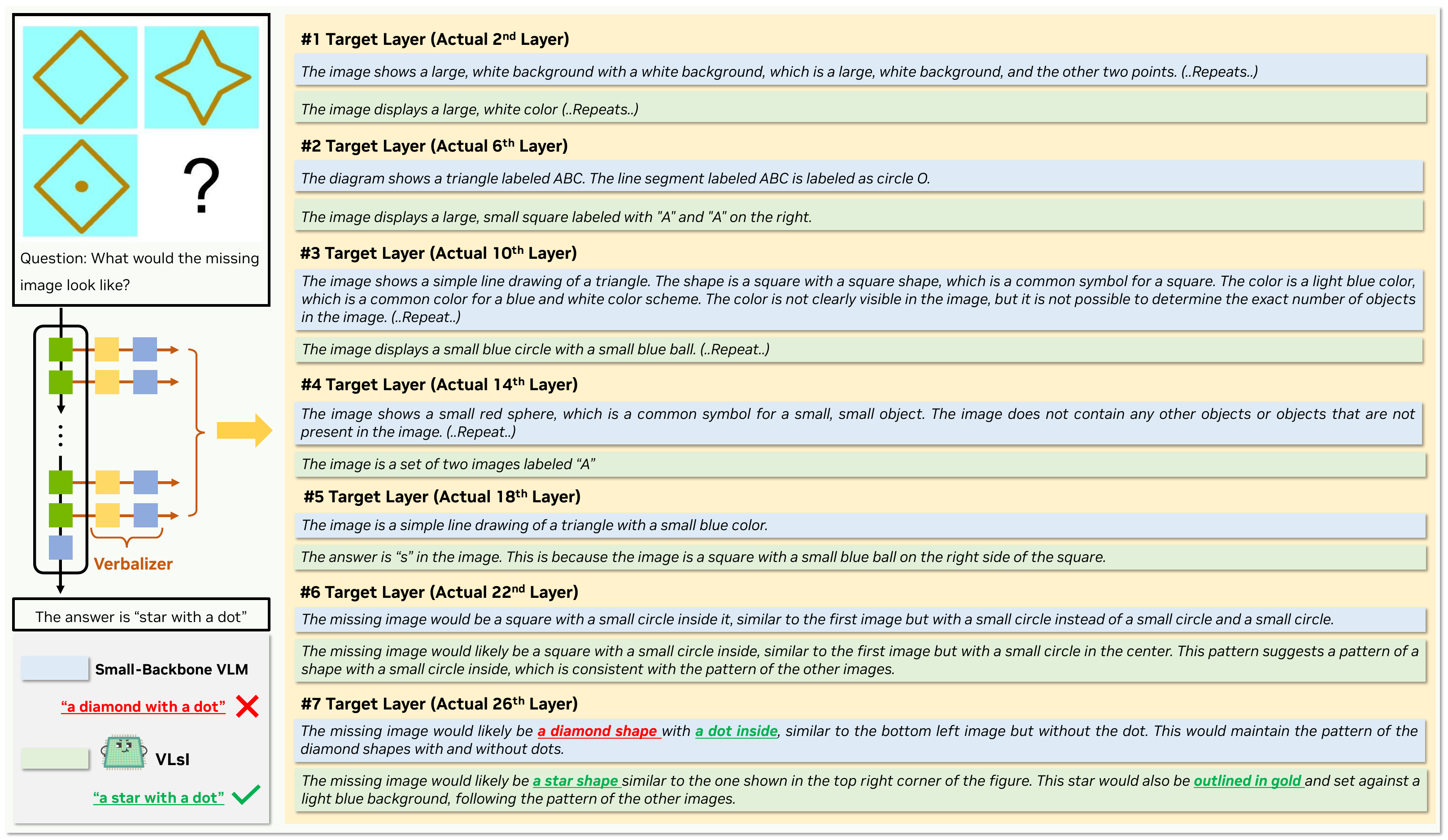}
    \caption{Example of verbalized outputs from each intermediate target layer in an alternative small-backbone VLM (without \vlsi VLsI enhancements) and the \vlsi VLsI. The visual question prompts VLM to predict the missing image in a sequence pattern. The outputs illustrate how each layer progressively interprets the visual cues, with \vlsi VLsI accurately identifying the answer as `a star with a dot' in the final layer, while the alternative small-backbone VLM incorrectly predicts `a diamond with a dot'. This demonstrates the improved interpretative capability of \vlsi VLsI through layer-wise, language-based distillation.
}
    \label{fig:3}
    \vspace{-5mm}
\end{figure*}
\section{VLsI: Verbalized Layers-to-Interactions}
\label{sec:method}

\paragraph{Overview of Model Architecture.} As illustrated in \cref{fig:2}, \vlsi VLsI comprises two main components: the backbone VLM and verbalizers. For the backbone VLM, we use Qwen2-VL~\cite{wang2024qwen2vl}, selected for its high performance on the OpenVLM-Leaderboard~\cite{2023opencompass}. The verbalizer consists of a simple FFN~\cite{vaswani2017attention} and the language head from the backbone VLM. To distinguish between FFN components, we refer to the FFN within the verbalizer as ``verb-FFN'' to avoid confusion with the FFN in the LLM transformer decoder block. This design is inspired by the recent speculative decoding paradigm~\cite{li2024eagle}, which demonstrates the effectiveness of using a smaller LLM constructed with a frozen word embedding and the language head of larger LLMs to emulate the performance of those larger models. Building on this insight, we incorporate the language head of the backbone VLM into its intermediate layers. Specifically, the verb-FFN is placed between each intermediate layer and the language head, introducing trainable parameters that allow effective projection into the language space. In other words, the verb-FFN enables outputs from the intermediate layers to project to the language space via the language head. In the following sections, we detail the three critical training stages: \textit{verbalization}, \textit{interaction}, and \textit{SFT}.

\subsection{Verbalization Step} In this step, we introduce a verbalizer for each target intermediate layer (see \cref{fig:2}(a)), allowing the outputs of these layers to be mapped to the natural language space. Each verbalizer comprises a verb-FFN (the yellow block in \cref{fig:2}(a)) and a language head (the blue block in \cref{fig:2}(a)), which is the same language head used in the corresponding backbone VLM. It is important to note that the outputs from intermediate layers are not directly translatable to the natural language space. To address this, we apply an additional verbalizer to process these intermediate layer outputs. To optimize the mapping from embeddings to natural language, we leverage instance pairs from visual instruction tuning datasets and apply an autoregressive loss to ensure the verbalized output aligns with the target response. Since the weights in both backbone VLMs are fixed, the gradient updates for each verbalizer at each layer remain independent. Our goal here is not for the intermediate layers to generate the correct response but rather to showcase their capacity to express verbal information given specific visual and text inputs. \cref{fig:3} presents the verbalization results, displaying the output of each layer along with its corresponding verbalizer. The results show a gradual improvement in reasoning and response accuracy as layers progress deeper. This verbalization enables our \vlsi VLsI to track the evolution of verbal responses in the natural language space, facilitating a clearer understanding of the `key developments' required to generate desired responses and thereby offering a more efficient distillation approach.

\definecolor{beaublue}{rgb}{0.74, 0.95, 0.85}
\newcommand{\cmark}{\ding{51}}%
\newcommand{\xmark}{\ding{55}}%

\begin{table*}[t!]
\vspace{-8mm}
\centering
\resizebox{\linewidth}{!}{
\renewcommand{\tabcolsep}{3mm}
\begin{tabular}{lccccccccccc}
\toprule
VLMs & QBench & AI2D & ChartQA & POPE & HallB & MME & MathVista & MMB & MMB$^{\text{CN}}$ & MM-Vet & MMMU\\
\midrule
LLaVA-NeXT-7B~\cite{liu2024llavanext}             & -    & -    & -    & 86.5 & -    & 1851 & 34.6 & 69.6 & 63.3 & 43.9 & 35.1 \\
LLaVA-NeXT-8B~\cite{liu2024llavanext}             & -    & 71.6 & 69.5 & -    & -    & 1972 & 37.5 & 72.1 & -    & -    & 41.7 \\
LLaVA-NeXT-13B~\cite{liu2024llavanext}            & -    & 70.0 & 62.2 & 86.7 & -    & 1892 & 35.1 & 70.0 & 68.5 & 47.3 & 35.9 \\
MM1-7B~\cite{mckinzie2024mm1}                     & -    & -    & -    & 86.6 & -    & 1858 & 35.9 & 72.3 & -    & 42.1 & 37.0  \\
MM1-MoE-7B$\times$32~\cite{mckinzie2024mm1}       & -    & -    & -    & 87.8 & -    & 1992 & 40.9 & 72.7 & -    & 45.2 & 40.9  \\
MiniGemini-HD-7B~\cite{li2024mini}                & -    & -    & -    & -    & -    & 1865 & 32.2 & 65.8 & -    & 41.3 & 36.8 \\
MiniGemini-HD-13B~\cite{li2024mini}               & -    & -    & -    & -    & -    & 1917 & 37.0 & 68.6 & -    & 50.5 & 37.3 \\
Cambrian-1-8B~\cite{tong2024cambrian}
            &  
            & 73.0 
            & 73.3 
            & - 
            & - 
            & - 
            & 49.0 
            & 75.9 
            & - 
            & - 
            & 42.7 \\ 
Cambrian-1-13B~\cite{tong2024cambrian}  
            &  
            & 73.6 
            & 73.8 
            & - 
            & - 
            & - 
            & 48.0 
            & 75.7 
            & - 
            & - 
            & 40.0 \\ 
Eagle-8B~\cite{shi2024eagle}  
            &  
            & 76.1 
            & 80.1 
            & - 
            & - 
            & - 
            & 52.7 
            & 75.9 
            & - 
            & - 
            & 43.8 \\ 
Eagle-13B~\cite{shi2024eagle}   
            &  
            & 74.0 
            & 77.6 
            & - 
            & - 
            & - 
            & 54.4 
            & 75.7 
            & - 
            & - 
            & 41.6 \\ 
VILA1.5-8B~\cite{lin2023vila}  
            & - 
            & -    
            & -
            & 85.6
            & -    
            & -
            & -
            & 75.3 
            & 69.9  
            & 43.2 
            & 38.6 \\
VILA1.5-13B~\cite{lin2023vila}  
            & - 
            & -    
            & -
            & 86.3
            & -    
            & -
            & -
            & 74.9 
            & 66.3 
            & 44.3 
            & 37.9 \\
VILA$^2$-8B~\cite{fang2024vila}  
            & -
            & -    
            & -
            & 86.7
            & -    
            & -
            & -
            & 76.6 
            & 71.7  
            & 50.0 
            & 38.3 \\
CogVLM2-8B~\cite{hong2024cogvlm2}
            & - 
            & 73.4 
            & 81.0 
            & - 
            & - 
            & 1870 
            & -  
            & 80.5 
            & - 
            & 60.4 
            & 44.3 \\ 
LLaVA-OneVision-7B~\cite{li2024llava}
            & - 
            & 81.4 
            & 80.0
            & - 
            & - 
            & 1998 
            & 63.2  
            & 80.8 
            & - 
            & 57.5 
            & 48.8 \\ 
InternVL2-8B~\cite{chen2023internvl}
            & - 
            & \underline{83.8} 
            & 83.3 
            & - 
            & - 
            & 2210 
            & 58.3  
            & 81.7 
            & \underline{81.2} 
            & 54.2 
            & 49.3 \\ 
MiniCPM-V2.5-8B~\cite{yao2024minicpm}
            & - 
            & - 
            & - 
            & - 
            & - 
            & 2025 
            & 54.3  
            & 77.2 
            & 74.2 
            & - 
            & 45.8 \\ 
MiniCPM-V2.6-8B~\cite{yao2024minicpm}
            & - 
            & - 
            & -
            & - 
            & - 
            & \textbf{2348} 
            & 60.6  
            & - 
            & - 
            & 60.0 
            & 49.8 \\ 
TroL-7B~\cite{lee2024trol}
            & 73.6 
            & 78.5 
            & 71.2 
            & 87.8 
            & 65.3 
            & 2308 
            & 51.8  
            & \underline{83.5} 
            & \underline{81.2} 
            & 54.7 
            & 49.9 \\ 
Phantom-7B~\cite{lee2024phantom}  
            & \underline{73.8}
            & 79.5    
            & \textbf{87.8}
            & 87.7
            & 65.4   
            & 2126
            & \underline{70.9}
            & 84.8 
            & 84.7  
            & \underline{70.8} 
            & 51.2    \\
Qwen2-VL-7B~\cite{wang2024qwen2vl}
            & \textbf{77.5} 
            & 77.5 
            & 83.0
            & \textbf{88.9} 
            & \underline{65.7} 
            & 2327 
            & 58.2 
            & 83.0 
            & 80.5 
            & 62.0 
            & \underline{54.1} \\ 
\midrule
\rowcolor{beaublue}
\vlsi
VLsI-7B
            & \textbf{77.5} 
            & \textbf{87.3} 
            & \underline{86.1} 
            & \underline{88.6} 
            & \textbf{74.2} 
            & \underline{2338} 
            & \textbf{74.7} 
            & \textbf{86.3} 
            & \textbf{85.5} 
            & \textbf{75.2} 
            & \textbf{69.3} \\ 
\bottomrule 
\end{tabular}
}
\caption{Evaluation of existing open-source VLMs and \vlsi VLsI on various vision-language benchmarks: QBench~\cite{wu2023q}, AI2D~\cite{kembhavi2016diagram}, ChartQA~\cite{masry2022chartqa}, POPE~\cite{li2023evaluating}, HallB~\cite{liu2023hallusionbench}, MME~\cite{fu2023mme}, MathVista~\cite{lu2023mathvista}, MMB~\cite{liu2023mmbench}, MMB$^{\text{CN}}$~\cite{liu2023mmbench}, MM-Vet~\cite{yu2023mm}, and MMMU~\cite{yue2023mmmu}. \textbf{Bold} and \underline{underline} indicate the top and second-best results, respectively.}
\label{tab:1}
\vspace{-3mm}
\end{table*}

\subsection{Interaction Step} After the \textit{verbalization step}, we proceed with a distillation process that leverages the natural language output from each intermediate layer. The main objective of this \textit{interaction step} is to establish an effective mapping between the layers of the large-backbone and small-backbone VLMs. This approach ensures that the small-backbone VLM mirrors the reasoning progression in the large one as layers deepen. Furthermore, because the computational requirements for each `key development' in generating the desired response vary, we propose an adaptive layer-matching strategy that dynamically aligns corresponding layers.

\paragraph{Extracting Verbal Information for Distillation.} To extract verbal information from intermediate layers, we use the vocabulary probabilities from the language head in the verbalizer based on input from the visual instruction training dataset. This method avoids the high computational cost of text generation (\textit{e.g.,} greedy or beam search~\cite{freitag-al-onaizan-2017-beam}), which would be prohibitively resource-intensive for each layer across both the large and small VLMs.

\paragraph{Layer Matching for Distilling Reasoning Progression.} To achieve effective layer matching, we develop a strategy to pair layers between the large- and small-backbone VLM, allowing the small model to learn the reasoning progression encoded in the large model. We define the number of target layers in the large and small VLMs as $t_{l}$ and $t_{s}$, respectively, with ranges $i_l=0^{\text{th}},1^{\text{st}},\cdots, (t_{l}-1)^{\text{th}}$ and $i_s=0^{\text{th}},1^{\text{st}},\cdots, (t_{s}-1)^{\text{th}}$, where $t_s \leq t_l$. Our matching strategy iterates over the $t_s$ layers in the small VLM, ensuring that each layer has a corresponding target layer in the large VLM. Note that, layers in the large-backbone VLM may not have corresponding layers in the small one. To maintain the reasoning progression as layers deepen, we employ a strategy that respects two key criteria: (i) Order Preservation—the matched layer $j$ (large-backbone VLM) of layer $i$ (small-backbone VLM) should be deeper than the matched layer $k$ of layer $i-1$, ensuring $j > k$; and (ii) Layer-wise Exploration (Multinomial Sampling)—to encourage novel and effective configurations for layer matching, we sample layers based on a distribution that is inversely proportional to the KL divergence between the verbal distributions of the matched layers in the large and small VLMs. Specifically, we compute this sampling distribution using a softmax of the scaled, negative KL divergence values, as summarized in \cref{alg:1}.

\begin{table*}[t!]
\vspace{-8mm}
\centering
\resizebox{\linewidth}{!}{
\renewcommand{\tabcolsep}{3mm}
\begin{tabular}{lccccccccccc}
\toprule
VLMs     & QBench & AI2D & ChartQA & POPE & HallB & MME & MathVista & MMB & MMB$^{\text{CN}}$ & MM-Vet & MMMU\\
\midrule
MiniCPM-2.4B~\cite{hu2024minicpm}                 & -    & 56.3 & -    & -    & -    & 1650 & 28.9 & 64.1 & 62.6 & 31.1 & -       \\
MiniCPM-V2-2.8B~\cite{hu2024minicpm}              & -    & 62.9 & -    & -    & -    & 1809 & 38.7 & 69.1 & 66.5 & 41.0 & -       \\
MM1-3B~\cite{mckinzie2024mm1}                     & - & -    & -    & 87.4 & -    & 1762 & 32.0 & 67.8 & -    & 43.7 & 33.9       \\
MM1-MoE-3B$\times$64~\cite{mckinzie2024mm1}       & - & -    & -    & 87.6 & -    & 1773 & 32.6 & 70.8 & -    & 42.2 & 38.6       \\
ALLaVA-3B~\cite{chen2024allava}                   & -    & -    & -    & -    & -    & 1623 & -    & 64.0 & -    & 32.2 & 35.3       \\
VILA1.5-3B~\cite{lin2023vila}                  & - & -    & -    & 85.3 & -    & -    & -    & 62.8 & 52.2 & 38.6 & 33.3    \\
InternVL2-4B~\cite{chen2023internvl}
            & - 
            & 78.9 
            & 81.5 
            & - 
            & - 
            & \textbf{2064} 
            & 58.6 
            & 78.6 
            & 73.9 
            & 51.0 
            & 34.3 \\ 
TroL-3.8B~\cite{lee2024trol}
            & 70.0 
            & 73.6  
            & 73.8 
            & 86.5 
            & \underline{62.2} 
            & 1980 
            & 55.1  
            & 79.2 
            & \underline{77.1}  
            & 51.1 
            & 37.5 \\ 
Phantom-3.8B~\cite{lee2024phantom} 
            & 70.3 
            & 71.7 
            & \textbf{87.3} 
            & 87.1 
            & 60.8 
            & \underline{2046} 
            & 60.6 
            & \underline{80.4} 
            & \underline{77.1} 
            & \underline{54.4} 
            & 39.2 \\ 
\midrule
DeepSeek-VL-1.3B~\cite{lu2024deepseek}            & -    & -    & -    & 87.6 & -    & -    & 31.1 & 64.6 & 62.9 & 34.8 & 32.2        \\
MobileVLM-1.7B~\cite{chu2023mobilevlm}            & - & -    & -    & 84.5 & -    & -    & -    & 53.2 & -    & -    & -        \\
MobileVLM-V2-1.7B~\cite{chu2024mobilevlm}         & - & -    & -    & 84.3 & -    & -    & -    & 57.7 & -    & -    & -        \\
MoE-LLaVA-1.8B$\times$4~\cite{lin2024moe}         & - & -    & -    & 87.0 & -    & -    & -    & 59.7 & -    & 25.3 & -       \\
Mini-Gemini-2B~\cite{li2024mini}                  & -    & -    & -    & -    & -    & 1653 & 29.4 & 59.8 & -    & -    & 31.7       \\
InternVL2-2B~\cite{chen2023internvl}
            & - 
            & \underline{74.1} 
            & 76.2 
            & - 
            & - 
            & 1877 
            & 46.3  
            & 73.2 
            & 70.9 
            & 39.5 
            & 34.3 \\ 
TroL-1.8B~\cite{lee2024trol}
            & 68.2 
            & 68.9 
            & 64.0 
            & \underline{88.6} 
            & 60.1  
            & 2038 
            & 45.4   
            & 76.1 
            & 74.1  
            & 45.1  
            & 35.2 \\ 
Phantom-1.8B~\cite{lee2024phantom} 
            & 69.1 
            & 62.3 
            & \underline{87.0} 
            & \textbf{89.6} 
            & \underline{62.2} 
            & 1885 
            & \underline{60.9} 
            & 76.6 
            & 75.1 
            & 54.1 
            & 40.6 \\ 
Qwen2-VL-2B~\cite{wang2024qwen2vl}
            & \underline{70.8} 
            & 60.2 
            & 73.5
            & 87.8 
            & 61.2 
            & 1872 
            & 43.0 
            & 74.9 
            & 73.5 
            & 49.5 
            & \underline{41.1} \\ 
\midrule
\rowcolor{beaublue}
\vlsi
VLsI-2B 
            & \textbf{72.3} 
            & \textbf{89.0} 
            & 85.8 
            & 87.9 
            & \textbf{70.0} 
            & 2022 
            & \textbf{68.4} 
            & \textbf{81.7} 
            & \textbf{78.8} 
            & \textbf{64.8} 
            & \textbf{51.4} \\ 
\bottomrule 
\end{tabular}
}
\caption{Comparison of smaller open-source VLMs and \vlsi VLsI on the same evaluation benchmarks as in Table~\ref{tab:1}.
}
\label{tab:2}
\vspace{-3mm}
\end{table*}

\setlength{\textfloatsep}{0pt}
\begin{algorithm}
\caption{Pseudo-Code for Interaction Loss}
\begin{algorithmic}[1]
    \STATE \textbf{Input:} $t_s$, $t_l$
    \STATE \textbf{Initialize:} loss: 0, $i_{l}^{*}$: 0, $\epsilon$: 1e-6, scale: 2
    \FOR{$i_s$ in $0\leq i_s<t_s$}
        \STATE \textit{kld-list} = []
        \FOR{$i_{l}$ in $i_{l}^{*}\leq i_{l}\leq t_{l}-t_{s}+i_s$ (Search Range)}
            \STATE \textit{kld-list}.append(\textit{compute-kld}($i_{s}$, $i_l$))
        \ENDFOR
        \STATE $T \gets \frac{\text{scale}}{\textit{kld-list.max}-\textit{kld-list.min}+\epsilon}$ \COMMENT{Temperature}
        \STATE $p\text{-list}\gets \text{Softmax}\left(-\textit{kld-list}/T\right)$
        \STATE $i_{l}^{*}\gets1+\text{Multinomial}(p\text{-list})$ \COMMENT{Sampling Index}
        \STATE loss $\gets$ loss $+$\textit{kld-list}[$i_{l}^{*}$]
    \ENDFOR
    \STATE \textbf{Return:} loss
\end{algorithmic}
\label{alg:1}
\end{algorithm}

\subsection{Supervised Finetuning (SFT) Step}

In the final stage of our distillation framework, we finetune the entire small-backbone VLM, including word embeddings, multi-head attention, FFN, and the language head on the visual instruction dataset in a supervised-learning manner. This \textit{SFT step} is inspired by pruning-based distillation methods~\cite{muralidharan2024compact, sreenivas2024llm}, which require additional training after pruning to counteract potential performance drops from structural changes. While our \textit{interaction step} does not alter the model’s structure, fully absorbing the rich information from the large VLM remains challenging for the small VLM within a single \textit{interaction step}. The SFT step allows the small VLM to better align and integrate the acquired knowledge across layers, enhancing response quality through autoregressive loss. We avoid optimizing autoregressive loss during the \textit{interaction step} (where KL divergence loss is optimized instead) as this can degrade performance, as demonstrated in our ablation study (\cref{sec:experi}). Therefore, to improve task-specific instruction-following responsiveness and accuracy, we incorporate the SFT step with autoregressive loss to further finetune the small-backbone VLM.
\section{Experiments}
\label{sec:experi}

\subsection{Implementation Details} We present four key technical components of \vlsi VLsI to ensure reproducibility: (a) the configuration of the backbone vision-language model, (b) the architecture of the verbalizer, (c) the training and inference configuration, and (d) the structure of the visual-instruction dataset, which includes a variety of capabilities crucial for effectively building \vlsi VLsI.

\paragraph{\textbf{(a) Configuration of the Backbone VLM.}} 
We select Qwen2-VL~\cite{wang2024qwen2vl} as our backbone VLM due to its flexible model scaling options, which include Qwen2-1.5B, Qwen2-7B, and Qwen2-72B~\citep{yang2024qwen2}. Importantly, the tokenizer’s vocabulary indices remain consistent across these model sizes, allowing for seamless integration without reordering the vocabulary. This structure enables us to focus on optimizing the LLM component, where Qwen2-1.5B and Qwen2-7B each contain 28 layers, and Qwen2-72B consists of 80 layers. For the vision encoder and projector, we adopt the same modules as those in Qwen2-VL: the vision encoder is a ViT model~\cite{dosovitskiy2020image} adapted from DFN~\cite{fang2023data} and enhanced with visually-adapted rotary positional embeddings~\cite{su2024roformer}. The vision projector comprises an MLP with two fully-connected layers interleaved with GELU activations~\cite{hendrycks2016gaussian}.

\begin{table*}[t!]
\vspace{-8mm}
\centering
\newcolumntype{g}{>{\columncolor{beaublue}}c}
\resizebox{\linewidth}{!}{
\renewcommand{\tabcolsep}{2mm}
\begin{tabular}{lcccccccgg}
\toprule
Benchmarks   & OmniFusion-7B & DeepSeek-VL-7B & MoVA-7B & Eagle-8B & CoLLaVO-7B & MoAI-7B & Meteor-7B &\vlsi VLsI-2B& \vlsi VLsI-7B \\
\midrule
MMB~\cite{liu2023mmbench} & 69.0 & 73.2  & 81.3    & 75.9  & \underline{83.0}& 79.3 & 82.9            & 81.7             & \textbf{86.3} \\
\cdashline{1-9}\noalign{\vskip 0.5ex}
MathVista~\cite{lu2023mathvista}& -  & -  & 44.3   & 52.7  & 57.6            & 56.2 & 53.4            & \underline{68.4} & \textbf{74.7}\\
\cdashline{1-9}\noalign{\vskip 0.5ex}
MM-Vet~\cite{yu2023mm}    & 39.4 & 41.5  & -       & -     & 40.3            & 43.7 & 57.3            & \underline{64.8} & \textbf{70.8}\\
\cdashline{1-9}\noalign{\vskip 0.5ex}
MMMU~\cite{yue2023mmmu}   & 36.6 & 36.6  & -       & 43.8  & 42.2            & 45.6 & 48.3            & \underline{51.4} & \textbf{69.3}\\
\bottomrule
\end{tabular}
}
\vspace{-2mm}
\caption*{(a) Validation of open-source VLMs with additional modules and projectors compared to \vlsi VLsI: OmniFusion~\cite{goncharova2024omnifusion}, DeepSeek-VL~\cite{lu2024deepseek}, MoVA~\cite{kar2024brave}, Eagle~\cite{shi2024eagle}, CoLLaVO~\cite{lee2024collavo}, MoAI~\cite{lee2024moai}, and Meteor~\cite{lee2024meteor}.
}

\vspace{5mm}
\vspace{-3mm}
\resizebox{\linewidth}{!}{
\renewcommand{\tabcolsep}{3mm}
\begin{tabular}{lcccccccccc}
\toprule
VLMs & MM-Vet & MM-Vet-v2 & MMMU & MMStar & AI2D & SEED-2-Plus & MathVista & BLINK & CV-Bench & LLaVA-Wilder \\
\midrule
LLaVA-NeXT-34B~\cite{liu2024llavanext}   & 50.7   & 50.9      & 48.8 & 51.6   & 78.9 & 65.9        & 40.4      & -     & -        & -            \\
VILA1.5-40B~\cite{lin2023vila}      & 51.2   & -         & 55.1 & 55.2   & 77.8 & -           & 49.5      & -     & -        & -            \\
Cambrian-34B~\cite{tong2024cambrian}     & 53.2   & -         & 50.4 & 54.2   & 79.5 & 65.1        & 50.3      & -     & 76.9     & -            \\
Molmo-72B~\cite{deitke2024molmo}        & 61.1   & -         & 52.8 & 63.3   & 83.4 & -           & 55.8      & -     & -        & -            \\
LLaVA-OV-72B~\cite{li2024llava}     & 60.6   & -         & 56.6 & 65.8   & 86.2 & -           & 68.4      & -     & -        & 72.0         \\
LLaMA-3.2-Vision & 64.1   & -         & 60.3 & 55.3   & 69.5 & 68.2        & 58.3      & 48.0  & -        & -            \\
Claude3.5-Sonnet~\cite{claude3series2024} & 66.0   & \textbf{71.8}      & 65.9 & 62.2   & 80.2 & 71.7        & 61.6      & 28.2  & -        & 83.1         \\
NVLM-D-72B~\cite{nvlm2024}       & 58.9   & -         & 60.8 & 63.7   & 80.1 & 68.4        & 63.9      & 48.0  & -        & -            \\
GPT-4V (0409)~\cite{chen2023sharegpt4v}    & 67.5   & 66.3      & 61.7 & 56.0   & 78.6 & 69.3        & 54.7      & 58.3  & 69.1     & 71.5         \\
Gemini-1.5-Pro   & 64.0   & 66.9      & 60.6 & 59.1   & 79.1 & 70.8        & 57.7      & 59.1  & -        & -            \\
InternVL2-76B~\cite{chen2023internvl}    & 64.4   & 68.4      & 58.3 & 67.1   & 87.6 & 70.0        & 65.6      & 57.5  & -        & -            \\
GPT-4o (0806)    & \underline{75.1}   & \underline{71.0}      & \textbf{69.9} & 64.7   & 84.7 & 70.8        & 62.7      & \textbf{64.7}  & -        & 85.9         \\
Qwen2-VL-72B~\cite{wang2024qwen2vl}     & 73.9   & 68.7         & 64.3 & 68.6   & \underline{88.3} & 72.3        & \underline{69.7}      & \underline{60.5}  & 74.3        & 84.1            \\
\midrule
TroL-1.8B~\cite{lee2024trol}     & 45.1   & -         & 35.2 & 45.5   & 68.9 & -           & 45.4      & -     & -        & -            \\
TroL-7B~\cite{lee2024trol}       & 54.7   & -         & 49.9 & 51.3   & 78.5 & -           & 51.8      & -     & -        & -            \\
Phantom-1.8B~\cite{lee2024phantom}     & 54.1   & 46.3      & 40.6 & 45.5   & 62.3 & 57.1        & 60.9      & 44.2  & 63.1     & 78.5         \\
Phantom-7B~\cite{lee2024phantom}       & 70.8   & 60.6      & 51.2 & 57.3   & 79.5 & 65.5        & 70.9      & 58.9  & 74.9     & 82.9         \\
\midrule
\rowcolor{beaublue}
\vlsi VLsI-2B  & 64.8            & 60.8      & 51.4 & \textbf{76.6}   & \textbf{89.0} & \textbf{81.1}        & 68.4      & 52.4  & \textbf{90.1}     & \underline{90.1}         \\
\rowcolor{beaublue}
\vlsi VLsI-7B    & \textbf{75.8}   & 70.0      & \underline{69.3} & \underline{73.6}   & 87.3 & \underline{74.9}        & \textbf{74.7}      & 59.7  & \underline{89.1}     & \textbf{92.0}         \\
\bottomrule
\end{tabular}
}
\vspace{-2mm}
\caption*{(b) Comparison of \vlsi VLsI with other open-source and closed-source VLMs on challenging benchmarks: MM-Vet~\cite{yu2023mm}, MM-Vet-v2~\cite{yu2024mm}, MMMU~\cite{yue2023mmmu}, MMStar~\cite{chen2024we}, AI2D~\cite{kembhavi2016diagram}, SEED-2-Plus~\cite{li2024seed}, MathVista~\cite{lu2023mathvista}, BLINK~\cite{fu2024blink}, CV-Bench~\cite{tong2024cambrian}, and LLaVA-Wilder~\cite{li2024llava}. This comparison includes models embedding additional knowledge~\cite{lee2024trol, lee2024phantom} and larger open/closed-source VLMs.
}

\caption{Detailed comparison of \vlsi VLsI with various open and closed-source VLMs on challenging evaluation benchmarks. Appendix A provides detailed descriptions of the evaluation benchmarks listed in Tables~\ref{tab:1} and \ref{tab:2}.
}
\label{tab:3}
\vspace{-3mm}
\end{table*}

\paragraph{\textbf{(b) Architecture of the Verbalizer.}} We design the verbalizer as a sequential feed-forward network (FFN), similar to the FFN typically used in transformer blocks~\cite{vaswani2017attention}, and as a language head of the backbone VLM. Conventionally, an FFN consists of three MLPs responsible for dimensional expansion, gating, and reduction. This configuration first expands the hidden dimension, then applies importance weighting to emphasize relevant features, and finally reduces the features back to the original hidden dimension. To enhance computational efficiency in verbalization and interaction, we opt to maintain a consistent hidden dimension throughout the process, foregoing the typical expansion and reduction steps. This streamlined FFN (verb-FFN) design reduces computational complexity while preserving overall performance. For intermediate target layers, we select $i_s$: 2$^{\text{nd}}$, 6$^{\text{th}}$, 10$^{\text{th}}$, ..., and 26$^{\text{th}}$ layers, and $i_t$: 2$^{\text{nd}}$, 6$^{\text{th}}$, 10$^{\text{th}}$, ..., and 78$^{\text{th}}$ layers.

\paragraph{\textbf{(c) Training and Inference Configuration.}} Training and inference for \vlsi VLsI are conducted on 8 NVIDIA A100 80GB GPUs. To enable efficient training, we apply LoRA~\citep{hu2021lora} to the LLM component, setting the rank and alpha parameters to 64. We use the AdamW optimizer~\citep{loshchilov2018decoupled} with a cosine annealing schedule, adjusting the learning rate from $1e^{-4}$ to $1e^{-6}$ over each training step. To handle large batch sizes effectively, we employ gradient accumulation with 16 steps and gradient checkpointing~\citep{sohoni2019low} to optimize memory usage. Specifically, batch configurations are four batches each for the 2B and 7B models and two batches for the 72B model, resulting in total gradient update counts of 512 (8×16×4) and 256 (8×16×2), respectively. For inference, we maintain the setup used in Qwen2-VL with a greedy search for text generation.

\paragraph{\textbf{(d) Visual-instruction dataset.}} Following the methodology in \cite{lee2024phantom}, we compiled a diverse dataset spanning a broad range of vision-language capabilities, totaling 2.9 million visual instruction tuning samples from various sources. This dataset includes foundational image understanding samples sourced from datasets such as ShareGPT4o-Images (57K)~\cite{sharegpt4o}, ShareGPT4V (755K)~\cite{chen2023sharegpt4v}, ALLaVA-VFLAN/Text (548K)~\cite{chen2024allava}, and MiniGemini (27K)~\cite{li2024mini}, which are focused on tasks like DocVQA~\cite{mathew2021docvqa}, ChartQA~\cite{masry2022chartqa}, DVQA~\cite{kafle2018dvqa}, and AI2D~\cite{kembhavi2016diagram}. To support scientific and mathematical reasoning, we incorporated samples from LLaVA-HD (116K)~\cite{zhang2024beyond}, enhancing datasets like ArXivQA~\cite{li2024multimodal} and TextbookQA~\cite{kembhavi2017you}. Additionally, we included document understanding samples from mPLUG-DocOwl1.5-Downstream/Reasoning (599K)~\cite{hu2024mplug}. For more general mathematical tasks, our dataset features samples from GLLaVA (177K)~\cite{gao2023g}, MathVision (3K)~\cite{wang2024measuring}, MathInstruct (262K)~\cite{yue2023mammoth}, and MathPlus (304K)~\cite{yue2024mammoth2}.

\begin{figure}[t!]
    \centering
    \includegraphics[width=0.3\textwidth]{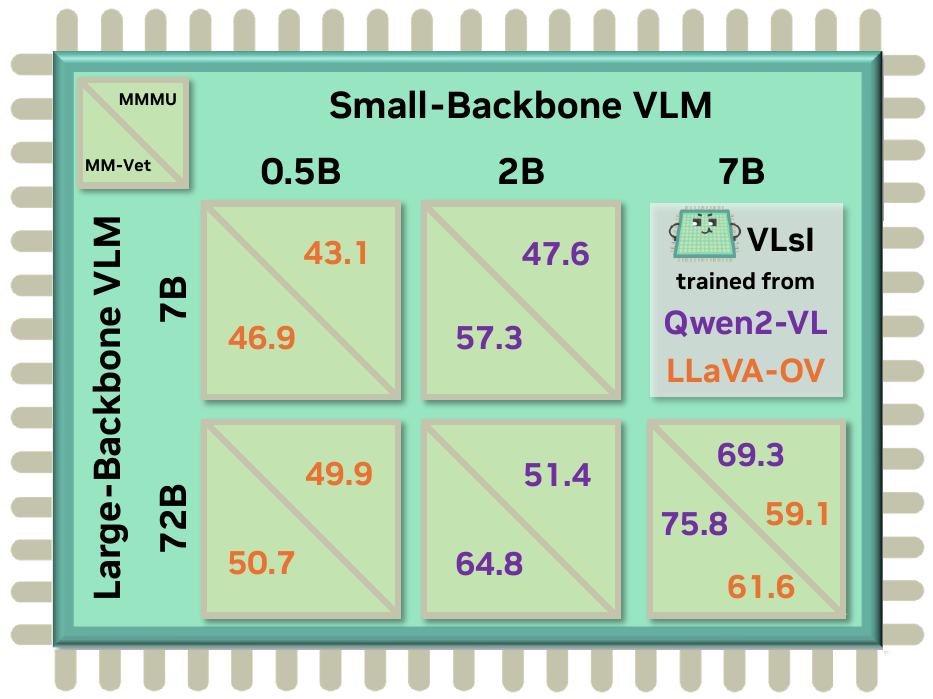}
    \vspace{-2mm}
    \caption{Comparison of performance on MM-Vet~\cite{yu2023mm} and MMMU~\cite{yue2023mmmu} across different model size combinations in large and small backbone VLMs. Each cell shows the evaluation results for various interaction configurations between 0.5B, 2B, and 7B small backbone VLMs trained with either Qwen2-VL~\cite{wang2024qwen2vl} or LLaVA-OV~\cite{li2024llava} as the large-backbone VLM.}
    \label{fig:4}
    \vspace{2mm}
\end{figure}

\begin{table*}[t!]
\vspace{-8mm}
\centering
\begin{minipage}[t]{0.33\linewidth}
\resizebox{\linewidth}{!}{
\renewcommand{\tabcolsep}{1.2mm}
\begin{tabular}{lccc}
\toprule
VLMs                   &MMB  &  MM-Vet & MMMU  \\
\midrule
LLaVA-OV-0.5B              &52.1 & 29.1  & 31.4 \\
\rowcolor{beaublue}
VLsI-0.5B (LLaVA-OV-72B)   &\textbf{72.5}& \textbf{50.7}  & \textbf{49.9}\\
\midrule
LLaVA-OV-7B                &80.8 & 57.5   &48.8 \\
\rowcolor{beaublue}
VLsI-7B (Qwen2-VL-72B)     &\textbf{86.3}  & \textbf{75.8}  &\textbf{69.3}  \\
\rowcolor{beaublue}
VLsI-7B (LLaVA-OV-72B)     & 86.1   & 61.6  & 59.1 \\
\midrule
LLaVA-OV-72B           &85.9  & 63.7   & 56.8\\
\bottomrule
\end{tabular}
}
\vspace{-3mm}
\caption*{(a) Backbone VLMs}

\vspace{2mm}
\resizebox{\linewidth}{!}{
\renewcommand{\tabcolsep}{2.2mm}
\begin{tabular}{llcccc}
\toprule
IL-Ops                & LL-Ops  &MMB & BLINK & MM-Vet & MMMU \\
\midrule
CE                    & \xmark  &79.2& 51.3	 & 64.5	  & 56.5\\
CE                    & CE      &77.8& 50.2	 & 63.2	  & 55.2 \\
CE                    & KLD     &81.0& 53.5	 & 67.2	  & 59.0 \\
L2                    & KLD     &81.5& 53.2	 & 66.8	  & 58.0 \\
\midrule
KLD                   & \xmark  &83.0& 55.0	 & 69.5	  & 61.0 \\
KLD                   & CE      &81.5& 54.3	 & 68.5	  & 59.8 \\
\rowcolor{beaublue}
KLD                   & KLD     &\textbf{86.3}&\textbf{59.7}  & \textbf{75.8}   & \textbf{69.3}\\
L2                    & KLD     &81.7&53.5	& 67.0	 & 58.3 \\
\bottomrule
\end{tabular}
}
\vspace{-3mm}
\caption*{(d) Operations for Intermediate/Last Layers}

\end{minipage}
\begin{minipage}[t]{0.33\linewidth}
\resizebox{\linewidth}{!}{
\renewcommand{\tabcolsep}{3.4mm}
\begin{tabular}{lccccc}
\toprule
VLMs          & SFT              &MMB  & BLINK & MM-Vet & MMMU\\
\midrule
Qwen2-VL-2B    & -               &74.9  & 41.4  & 49.5  &41.1\\
VLsI-2B        & \xmark          &73.2	& 40.1	& 47.9	&39.8\\
VLsI-2B (Beam) & \xmark          &77.6	& 48.3	& 58.1	&45.9\\
\rowcolor{beaublue}
VLsI-2B        & \cmark          &81.7  & 52.4  & 64.8  &51.4\\
\midrule
Qwen2-VL-7B    & -               &83.0  & 50.8  & 62.0  &54.1\\
VLsI-7B        & \xmark          &82.1	& 49.6	& 60.5	&52.9\\
VLsI-7B (Beam) & \xmark          &85.2	& 56.3	& 70.9	&63.5\\
\rowcolor{beaublue}
VLsI-7B        & \cmark          &86.3  & 59.7  & \textbf{75.8}  &\textbf{69.3}\\
\midrule
Qwen2-VL-72B   & -               &\textbf{86.5}  & \textbf{60.5}  & 73.9  &64.3\\
\bottomrule
\end{tabular}
}
\vspace{-3mm}
\caption*{(b) Use of \textit{SFT Step}}

\vspace{1.5mm}
\resizebox{\linewidth}{!}{
\renewcommand{\tabcolsep}{1.3mm}
\begin{tabular}{lcccc}
\toprule
VLMs                            & MMB & BLINK     & MM-Vet & MMMU\\
\midrule
Random Index                    & 77.0& 50.0	  & 62.0   & 52.0 \\
Uniform Index                   & 79.5& 51.9	  & 66.5   & 55.3\\
Bottom-1 Index                  & 81.2&	53.8	  & 67.8   & 57.7\\
Bottom-3 Index                  & 81.5& 54.0	  & 68.0   & 58.0\\
\midrule
\rowcolor{beaublue}
Multinomial Sampling            & 82.0& 54.5	  & 68.5   & 58.5\\
\rowcolor{beaublue}
+Search Range                   & 83.5& 55.5	  & 69.8   & 60.0\\
\rowcolor{beaublue}
+Order Preservation             & 86.0& 59.2	  & 75.2   & 68.5\\
\rowcolor{beaublue}
+Adaptive Temperature           &\textbf{86.3} & \textbf{59.7}  & \textbf{75.8}   & \textbf{69.3}\\
\bottomrule
\end{tabular}
}
\vspace{-3mm}
\caption*{(e) Components in Matching Strategy}

\end{minipage}
\begin{minipage}[t]{0.33\linewidth}
\centering
\resizebox{\linewidth}{!}{
\renewcommand{\tabcolsep}{0.8mm}
\begin{tabular}{lcccc}
\toprule
RS Training Percent.  & MMB  & BLINK & MM-Vet& MMMU \\
\midrule
Qwen2-VL-7B             & 80.5 & 50.8  & 62.0  & 54.1\\
+50\%                   & 81.0 & 51.5  & 62.8  & 54.6\\
+100\%                  & 81.8 & 52.3  & 63.7  & 55.2\\
\midrule
VLsI-7B                 & 82.1 & 49.6  & 60.5 & 52.9\\
+50\%                   & 85.4 & 56.0  & 70.0 & 62.1\\
\rowcolor{beaublue}
+100\%                  & \textbf{86.3} & \textbf{59.7}  & \textbf{75.8} & \textbf{69.3}\\
\bottomrule
\end{tabular}
}
\vspace{-3mm}
\caption*{(c) Percentage of Training Iterations in RS}
\vspace{1.6mm}
\resizebox{\linewidth}{!}{
\renewcommand{\tabcolsep}{1.2mm}
\begin{tabular}{lrcccc}
\toprule
Structure                   & Params& MMB   & BLINK & MM-Vet & MMMU \\
\midrule
Decoder$\times 2$           & 3.3B  & 85.5  & 59.1	& 76.2	 & 68.6\\
Decoder                     & 1.6B  & 85.7  & 59.0	& 76.0	 & 68.4\\
FFN$\times 2$               & 2.9B  & 86.3  & 59.2	& 75.9	 & 69.3\\
FFN                         & 1.4B  & 86.4  & 59.4	& 75.7	 & 69.2\\
verb-FFN$\times 2$          & 539M  & 85.8  & 59.9  & 75.8   & 69.3\\
\rowcolor{beaublue}
verb-FFN                    & 269M  & 86.3  & 59.7  & 75.8   & 69.3\\
MLP$\times 2$               & 180M  & 84.2	& 57.5	& 74.1	 & 67.0\\
MLP                         & 90M   & 83.8	& 57.0	& 73.5	 & 66.7\\
\bottomrule
\end{tabular}
}
\vspace{-3mm}
\caption*{(f) Verbalizer Architecture}
\end{minipage}
\vspace{-2mm}
\caption{Ablation studies examining the six main factors influencing the effectiveness of \vlsi VLsI.
}
\label{tab:4}
\end{table*}

\begin{figure*}[t!]
    \centering
    \vspace{-3mm}
    \includegraphics[width=\textwidth]{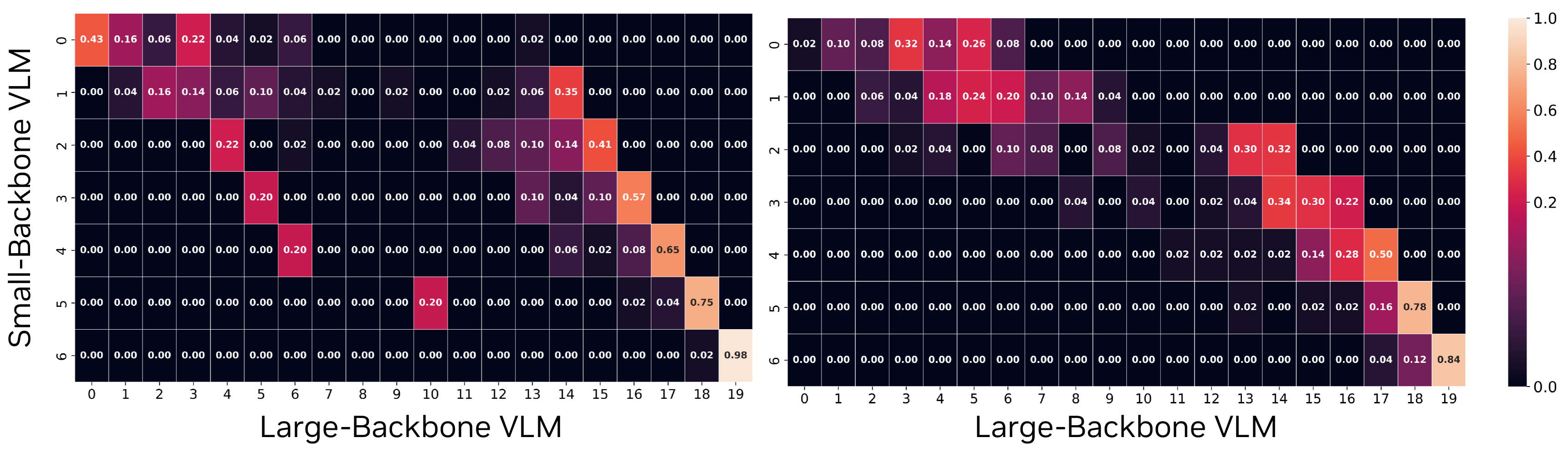}
    \vspace{-7mm}
    \caption{
    Distribution changes of the matched indices between small-backbone and large-backbone VLMs at the interaction step. The left figure shows the distribution at the beginning of training, while the right figure shows it at the end.}
    \label{fig:5}
    \vspace{-3mm}
\end{figure*}

\subsection{Verbalization on \vlsi VLsI}

\cref{fig:3} illustrates the verbal responses generated at each intermediate layer in small-backbone VLM and \vlsi VLsI. Using the verbalized outputs to trace each layer's interpretive progression, this comparison highlights how both models gradually enhance understanding across layers. At the shallower layers, both models generate basic descriptions, focusing on large, simple shapes and colors. However, as \vlsi VLsI progresses to mid-level layers, it begins to recognize and articulate more complex visual structures, such as labeled shapes and their relative positions. In contrast, the small-backbone VLM’s verbal responses remain relatively vague or repetitive, often lacking in specific relational details. By the deeper layers, \vlsi VLsI demonstrates a clear advantage: its verbalizations shift towards identifying the correct pattern, explicitly referring to shapes and colors in alignment with the target response: ``star with a dot''. Meanwhile, the small-backbone VLM incorrectly predicts the missing image as a ``diamond with a dot'', failing to capture the specific pattern. This example underscores the effectiveness of \vlsi VLsI’s layer-wise verbalization, where each stage of verbal responses helps the small-backbone VLM align with the larger one. Note that, Appendix B shows the loss changes during verbalization. Additional examples of \vlsi VLsI’s verbalization are available in Appendix C, highlighting its capacity to interpret layer-wise verbal responses.

\subsection{Comparison on Evaluation Benchmarks} As shown in \cref{tab:1} and \cref{tab:2}, \vlsi VLsI achieves higher performance while maintaining an efficient model scale. Furthermore, \cref{tab:3} compares \vlsi VLsI’s performance with (a) open-source VLMs that incorporate multiple vision encoders, computer vision models, and additional rationale projector; (b) VLMs with modified architectures~\cite{lee2024trol, lee2024phantom} and various larger open- and proprietary closed-source VLMs. To evaluate where the effectiveness comes from, we analyze six key factors detailed in \cref{fig:4} and \cref{tab:4}. Appendix D explains the detailed settings for ablation studies. These results indicate that (1) utilizing a more capable large-backbone VLM provides substantial performance benefits, suggesting that the choice of backbone significantly impacts the transfer of knowledge; (2) using larger-backbone VLM gets benefits; (3) KL divergence is more effective during interaction step than cross-entropy, and simultaneously using last-layer distillation boosts performances; and (4) additional SFT step is crucial for further performance gains, consistent with findings in~\cite{muralidharan2024compact, sreenivas2024llm}. Appendix E represents \vlsi VLsI's text generation quality.

\subsection{Discussion and Limitations} For \cref{fig:3}, \vlsi VLsI's verbalizer is not re-trained but just utilized by the trained verbalizer of the small-backbone VLM. Interestingly, this verbalizer also works well at \vlsi VLsI, demonstrating flexibility and indicating that it may serve as a practical medium of interpretability. Additionally, \cref{fig:5} illustrates that, as the interaction step progresses, the small-backbone VLM gradually tries to learn about deeper layers' responses of the large-backbone VLM, which can be considered accelerating the process of reaching an answer. While \vlsi VLsI is highly effective, the large and small-backbone VLMs must share the same tokenizer and token index order when constructing VLsI. We will explore more general ways that accommodate different tokenizers and token index orders, potentially expanding \vlsi VLsI’s applicability and scalability.
\section{Conclusion}
\label{sec:conclusion}

We present a new VLM family with 2B and 7B model sizes, \vlsi VLsI: Verbalized Layers-to-Interactions, designed to build high-performing yet efficient-scale VLMs. This is accomplished by leveraging natural language-based distillation to transfer knowledge from large to small VLMs. We show \vlsi VLsI achieves strong vision-language performances, suggesting that natural language is an important key in transferring knowledge not only for humans but also for AI. We hope to keep progress in further utilization of natural language and larger VLMs.

\clearpage
{
    \small
    \bibliographystyle{ieeenat_fullname}
    \bibliography{main}
}

\clearpage
\setcounter{page}{1}
\onecolumn

\appendix

\section{Description of Numeorus Evaluation Benchmarks}
\label{app:A}
\begin{itemize}    
    \item \textbf{QBench}~\citep{wu2023q} is a comprehensive benchmark designed to evaluate the low-level visual abilities of multimodal large language models (MLLMs), focusing on perception, description, and quality assessment of visual attributes. It introduces datasets like LLVisionQA for diverse low-level attribute queries, LLDescribe for detailed expert-crafted image descriptions, and a unified softmax-based strategy for quantifiable image quality assessment. Q-Bench highlights that while MLLMs exhibit preliminary capabilities in handling low-level visual tasks, their outputs remain inconsistent and imprecise, emphasizing the need for further advancements to align with human perception and achieve general-purpose applications.
    
    \item \textbf{AI2D}~\citep{kembhavi2016diagram} is a benchmark dataset developed to study diagram interpretation and reasoning, focusing on identifying diagram structures and semantic relationships. It introduces Diagram Parse Graphs (DPG), a graph-based representation that encodes the syntactic and semantic structure of diagrams. The dataset contains over 5,000 grade-school science diagrams with exhaustive annotations of constituents and relationships, as well as 15,000 multiple-choice questions for diagram-based reasoning tasks.
    
    \item \textbf{ChartQA}~\citep{masry2022chartqa} is a large-scale benchmark designed to assess question-answering systems' ability to reason logically and visually about data visualizations like bar, line, and pie charts. It includes 9,608 human-authored questions and 23,111 machine-generated questions, focusing on complex reasoning tasks involving mathematical operations, visual attributes, and multi-step logical inferences. By utilizing both extracted data tables and visual features, the benchmark highlights challenges in handling real-world charts and emphasizes the gap in models' ability to process intricate visual and logical questions compared to human understanding.

    \item \textbf{SEED-Bench-2-Plus}~\citep{li2024seed} is a comprehensive benchmark designed to evaluate Multimodal Large Language Models (MLLMs) on their ability to comprehend and reason about text-rich visual content across three categories: Charts, Maps, and Webs, covering 63 diverse data types. It includes 2.3K meticulously crafted multiple-choice questions with human-verified answers, simulating real-world scenarios that combine complex text and visual data.
    
    \item \textbf{POPE}~\citep{li2023evaluating} is a polling-based evaluation framework designed to assess object hallucination in Large Vision-Language Models (LVLMs). It formulates hallucination detection as a binary classification task using simple yes/no questions (e.g., ``Is there a chair in the image?'') to probe LVLMs. Unlike previous methods, POPE offers a stable and flexible approach by avoiding dependence on lengthy captions or complex parsing rules. It introduces three object sampling strategies—Random, Popular, and Adversarial—to explore hallucination patterns in models.
    
    \item \textbf{HallusionBench (HallB)}~\citep{liu2023hallusionbench} is an advanced diagnostic benchmark designed to evaluate and analyze the failure modes of Large Vision-Language Models (LVLMs) in handling both language hallucinations and visual illusions. Featuring 346 images and 1,129 human-crafted visual-question pairs, it tests models like GPT-4V and LLaVA-1.5 using unique control pairs and human-edited images to assess logical consistency, perception, and reasoning. Results highlight persistent challenges, with top models achieving only 31.42\% accuracy, revealing their over-reliance on parametric memory, susceptibility to simple manipulations, and struggles with geometry, math, and temporal reasoning.
    
    \item \textbf{MME}~\citep{fu2023mme} is a comprehensive benchmark designed to evaluate Multimodal Large Language Models (MLLMs) across perception and cognition abilities with 14 subtasks. The benchmark includes tasks like object recognition, OCR, commonsense reasoning, and numerical calculation, using manually curated instruction-answer pairs to ensure fairness and avoid data leakage. MME emphasizes concise instructions for consistency and quantitative assessment, highlighting that current MLLMs, despite their progress, face challenges such as instruction-following errors, limited perception and reasoning, and hallucinations.
    
    \item \textbf{MathVista}~\citep{lu2023mathvista} is a benchmark designed to evaluate the mathematical reasoning abilities of foundation models in visual contexts. It comprises 6,141 examples sourced from 31 datasets, including three newly created datasets—IQTest, FunctionQA, and PaperQA—tailored to assess logical, algebraic, and scientific reasoning in visual settings. MathVista emphasizes diverse visual contexts, such as diagrams, charts, and academic figures, and covers seven types of reasoning across five core tasks.
    
    \item \textbf{MMB, MMB-Chinese (MMB$^{\text{CN}}$)}~\citep{liu2023mmbench} is a multilingual benchmark designed to evaluate the multimodal capabilities of vision-language models (VLMs) across 20 fine-grained abilities, including perception, reasoning, and relation understanding. It features over 3,000 high-quality multiple-choice questions in English and Chinese, enabling comparative analysis in a bilingual context. MMBench introduces novel evaluation strategies like CircularEval, which enhances robustness by testing models across shuffled choices, and employs GPT-4 for accurate choice extraction.
    
    \item \textbf{MM-Vet}~\citep{yu2023mm} is a benchmark designed to evaluate the integrated vision-language capabilities of Large Multimodal Models (LMMs). It defines six core capabilities—recognition, OCR, knowledge, language generation, spatial awareness, and math—and examines their combinations across 16 emergent multimodal tasks, such as explaining memes, solving spatial math problems, and summarizing visual data. The benchmark introduces an LLM-based evaluator to assess open-ended model outputs consistently, focusing on both accuracy and quality.

    \item \textbf{MM-Vet-v2}~\citep{yu2024mm} builds upon the original MM-Vet benchmark by introducing a new core capability, image-text sequence understanding, to evaluate large multimodal models (LMMs) on processing arbitrarily interleaved sequences of images and text. With an expanded dataset of 517 high-quality evaluation samples and tasks requiring combinations of seven core capabilities, it assesses advanced real-world scenarios like temporal reasoning, spatial understanding, and multimodal comparisons.

    \item \textbf{MMMU~\cite{yue2023mmmu}} is a benchmark designed to evaluate large multimodal models on 11550 college-level problems requiring expert knowledge and reasoning across six disciplines: Art, Business, Science, Medicine, Humanities, and Engineering. Spanning 30 subjects and incorporating 30 diverse image types like charts, medical scans, and diagrams, it challenges models to integrate complex text and image inputs while applying domain-specific knowledge. MMMU sets a high standard for advancing multimodal AI and plays a crucial role in developing Expert AGI.
    
    \item \textbf{MMStar}~\citep{chen2024we} is a vision-critical multimodal benchmark consisting of 1,500 meticulously curated samples designed to evaluate large vision-language models (LVLMs) across six core capabilities and 18 specific axes. By addressing two key issues in existing benchmarks—unnecessary reliance on textual knowledge and unintentional data leakage—MMStar ensures each sample requires genuine visual reasoning and minimal data recall. Incorporating metrics for multimodal gain and data leakage, it provides a robust platform for assessing the true multimodal capacities of LVLMs.

    \item \textbf{BLINK}~\citep{fu2024blink} is a benchmark designed to evaluate the core visual perception abilities of multimodal large language models (MLLMs) across 14 tasks, such as depth estimation, visual correspondence, and spatial reasoning, inspired by classical computer vision problems. With 3,807 multiple-choice questions derived from 7,300 images, BLINK focuses on tasks that humans can solve ``within a blink'' but remain challenging for models, as even advanced models like GPT-4V achieve only 51.26\% accuracy compared to 95.7\% for humans. It highlights the gap in nuanced visual perception and suggests integrating specialized vision models as a pathway for improving MLLMs' performance.

    \item \textbf{CV-Bench}~\citep{tong2024cambrian} is a vision-centric benchmark introduced to evaluate the fundamental 2D and 3D visual understanding capabilities of Multimodal Large Language Models (MLLMs). With 2,638 manually inspected examples sourced from datasets like ADE20K, COCO, and Omni3D, it tests tasks such as spatial relationships, object counting, depth ordering, and relative distances. By transforming traditional vision benchmarks into VQA format, CV-Bench ensures robust assessment of models' abilities in multimodal contexts. It addresses gaps in existing benchmarks by offering significantly more samples, better diversity, and a stronger focus on visual grounding, making it a critical tool for advancing multimodal AI systems.

    \item \textbf{LLaVA-Wilder}~\citep{zhang2024lmms} is a dataset designed to evaluate large multimodal models (LMMs) in real-world scenarios. It comprises 128 image-text pairs, each featuring an image accompanied by a question and a detailed answer. The dataset includes a variety of images, such as indoor and outdoor scenes, memes, paintings, and sketches, to assess models' generalization capabilities across diverse domains. By providing this resource, LLaVA-Bench-Wilder aims to facilitate the development and benchmarking of LMMs, ensuring their robustness and effectiveness in handling complex, real-world visual tasks.
\end{itemize}

\section{Verbalization Loss Changes}
\label{sec:appB}
\begin{table}[h!]
\vspace{-4mm}
\centering
\resizebox{\linewidth}{!}{
\renewcommand{\tabcolsep}{10mm}
\begin{tabular}{ccccccc}
\toprule
Target Layer Index  & 10\% & 30\% & 50\% & 70\% & 90\% & 100\% \\
\midrule
\#1 & 14.28 & 11.73 &  9.87 &  6.78 &  3.54 &  3.30 \\
\#2 & 12.60 & 10.48 &  8.04 &  5.98 &  3.10 &  2.90 \\
\#3 & 11.22 &  9.32 &  7.02 &  5.05 &  2.63 &  2.37 \\
\#4 & 10.48 &  8.11 &  6.41 &  4.57 &  2.35 &  2.18 \\
\#5 &  9.34 &  7.55 &  5.81 &  3.54 &  1.82 &  1.79 \\
\#6 &  8.16 &  6.18 &  4.27 &  2.09 &  1.56 &  1.47 \\
\#7 &  7.39 &  3.85 &  2.02 &  1.03 &  0.85 &  0.79 \\
\bottomrule
\end{tabular}
}
\caption*{Showing loss changes during the verbalization step (\% indicates training progress)}
\vspace{-4mm}
\end{table}

\clearpage
\section{VLsI's Verbalization Examples}
\label{sec:appC}
\begin{figure}[h!]
    \centering
    \includegraphics[width=\textwidth]{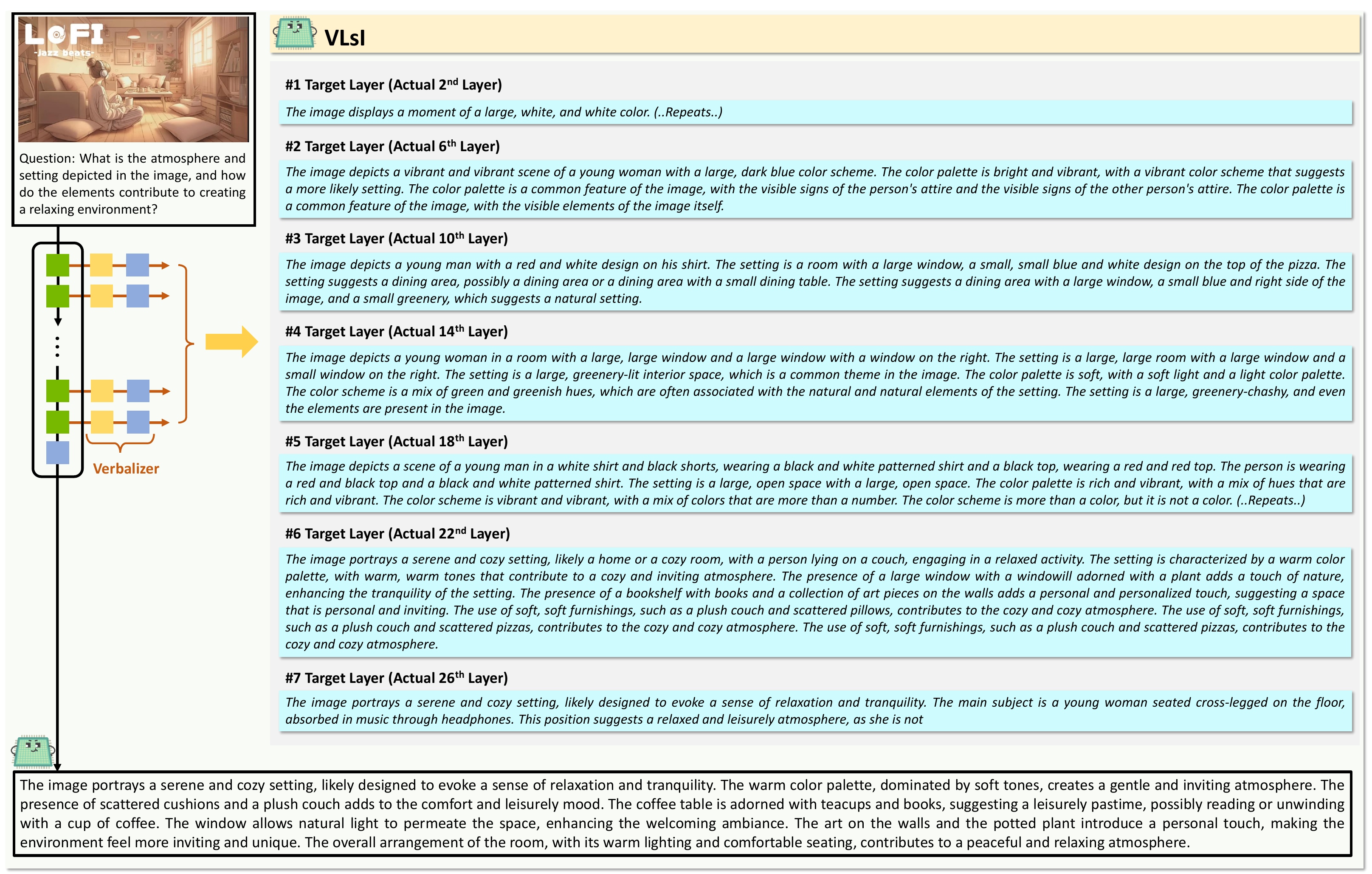}
\end{figure}

\begin{figure}[h!]
    \centering
    \includegraphics[width=\textwidth]{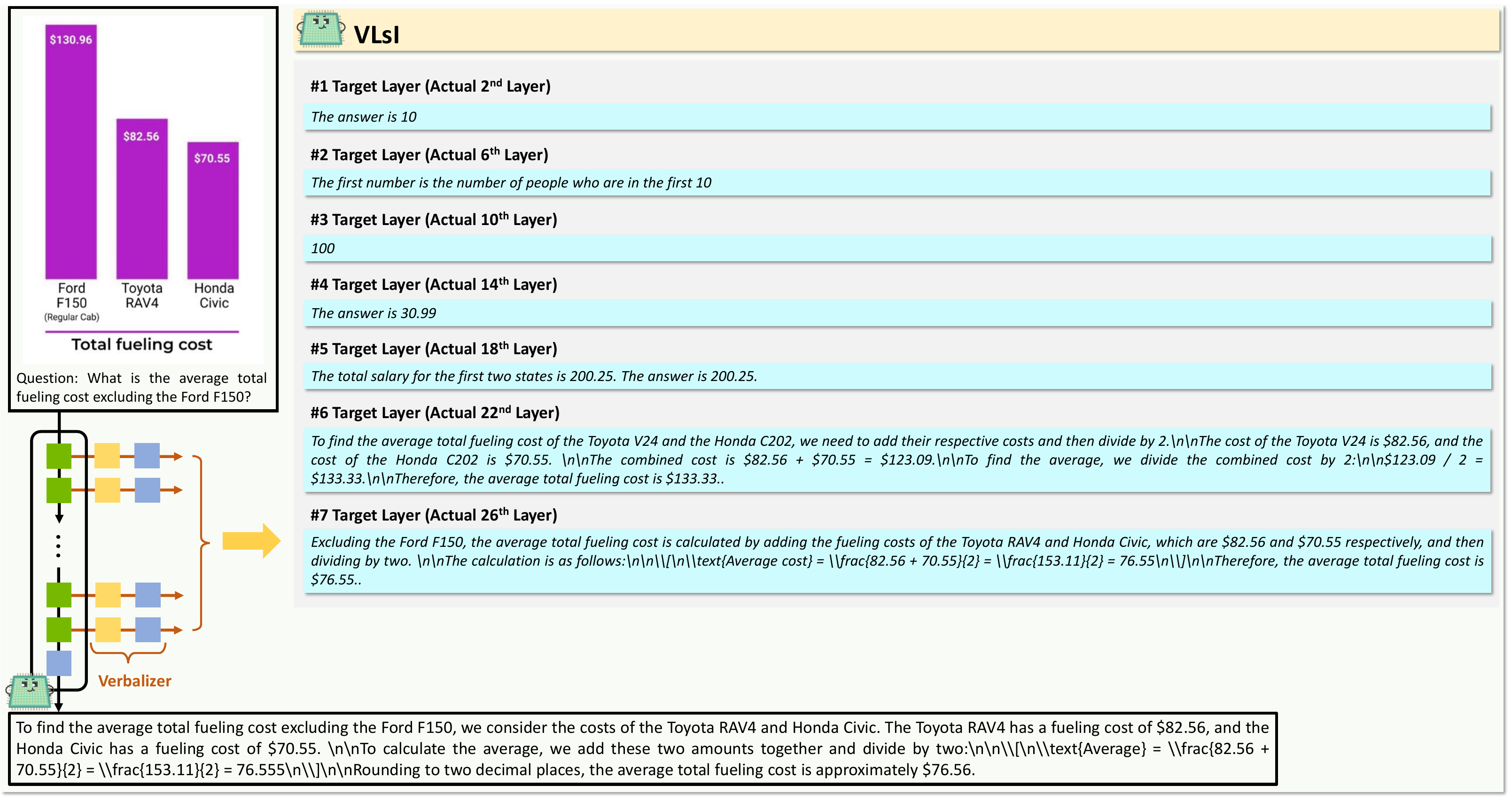}
\end{figure}

\begin{figure}[h!]
    \centering
    \includegraphics[width=\textwidth]{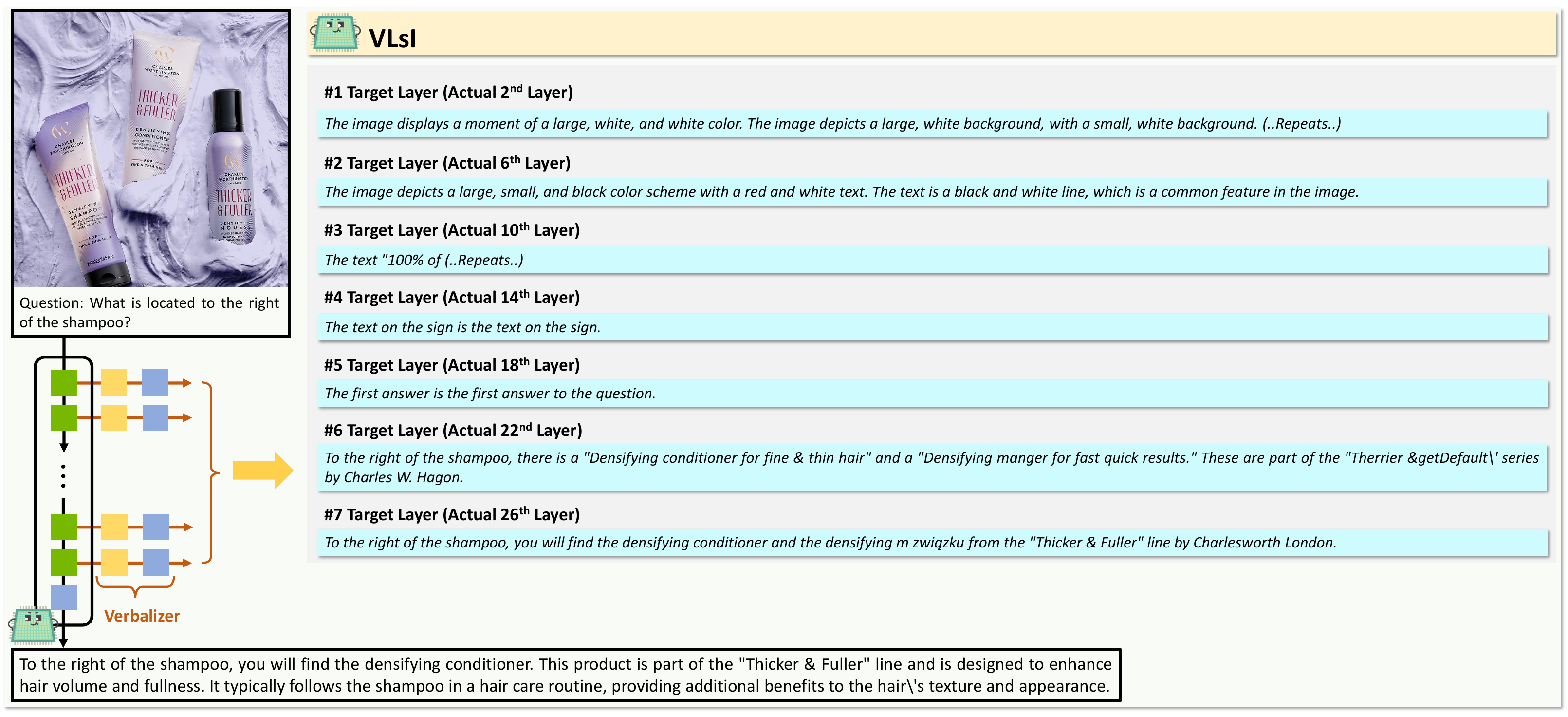}
\end{figure}

\begin{figure}[h!]
    \centering
    \includegraphics[width=\textwidth]{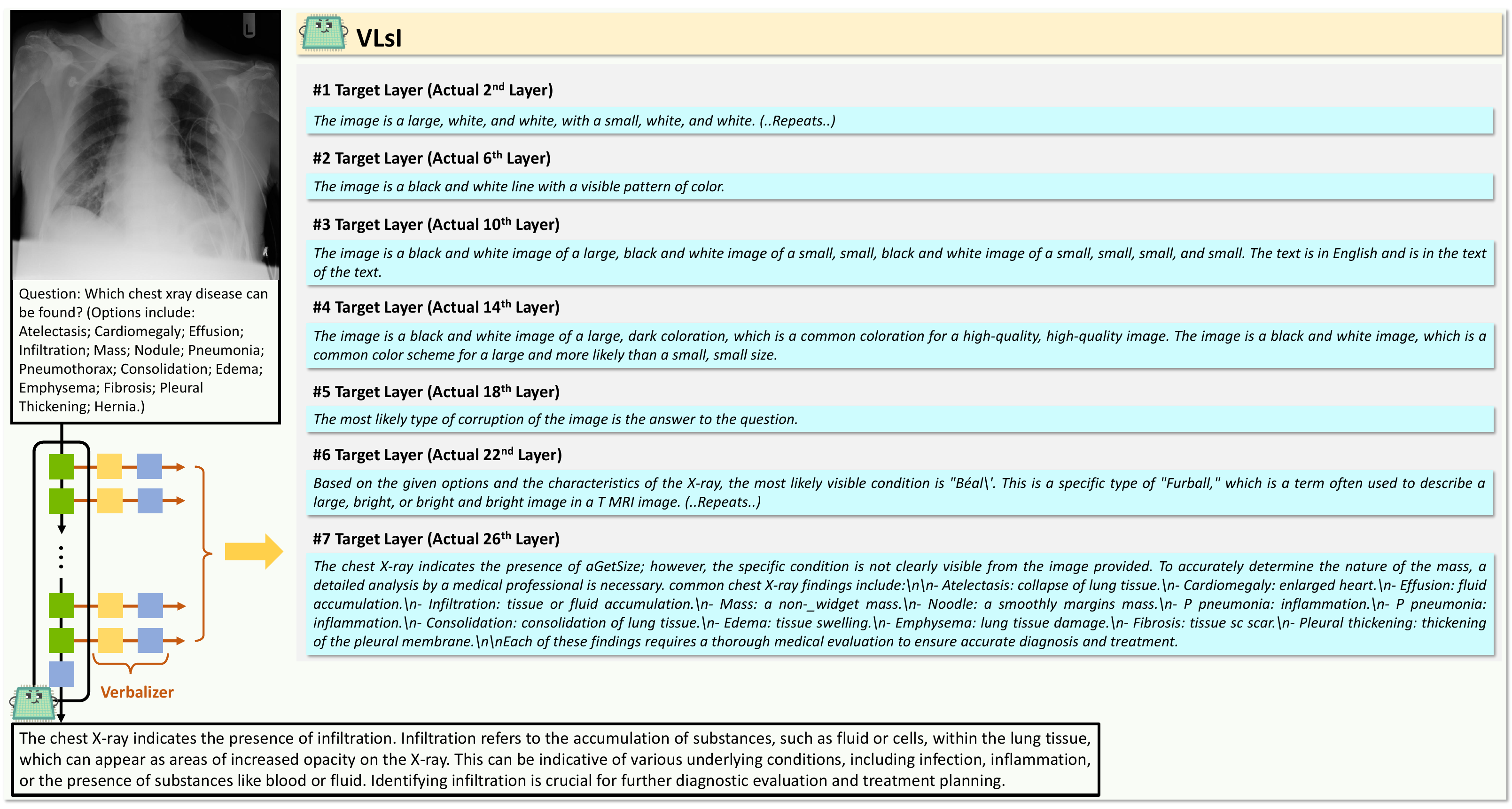}
\end{figure}

\begin{figure}[h!]
    \centering
    \includegraphics[width=\textwidth]{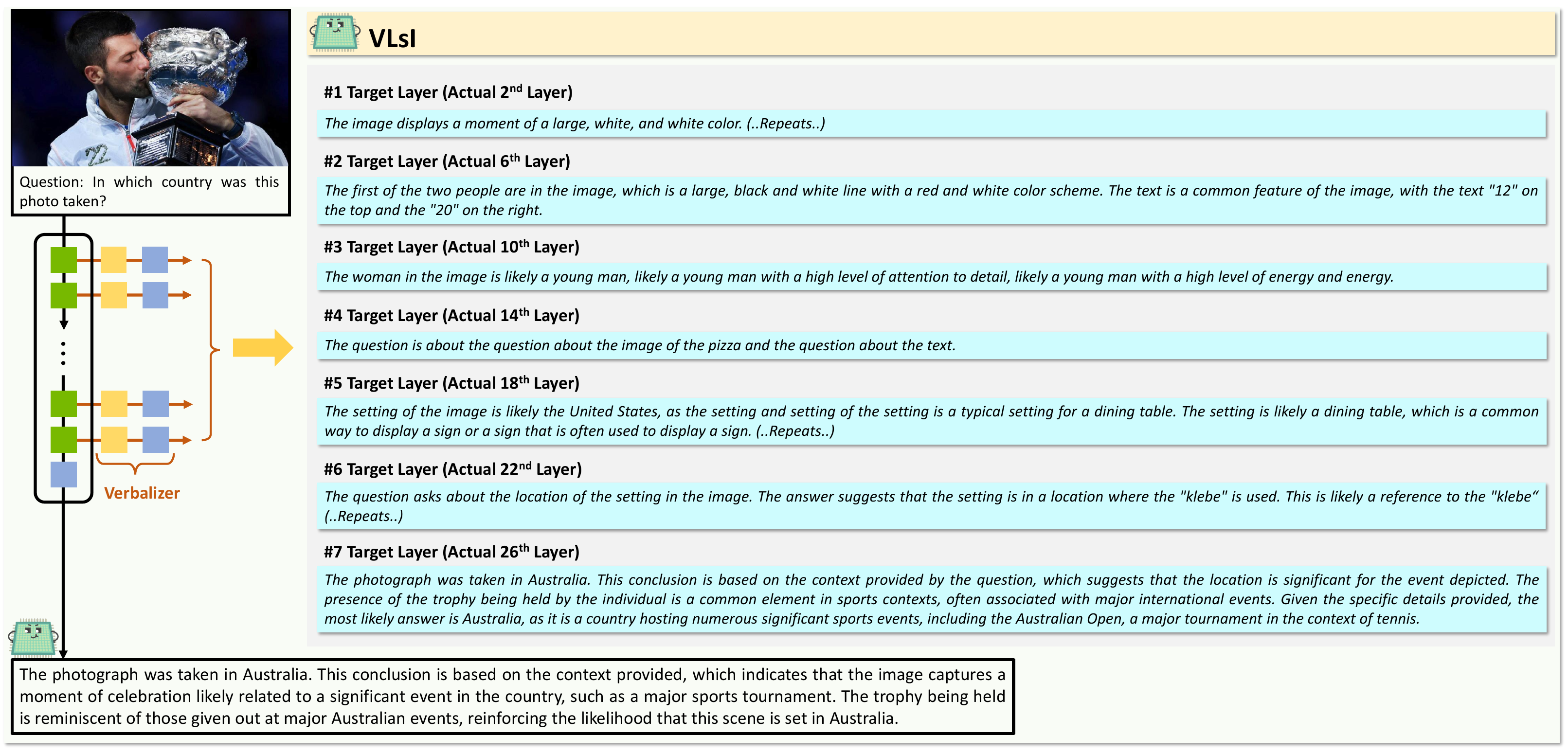}
\end{figure}

\begin{figure}[h!]
    \centering
    \includegraphics[width=\textwidth]{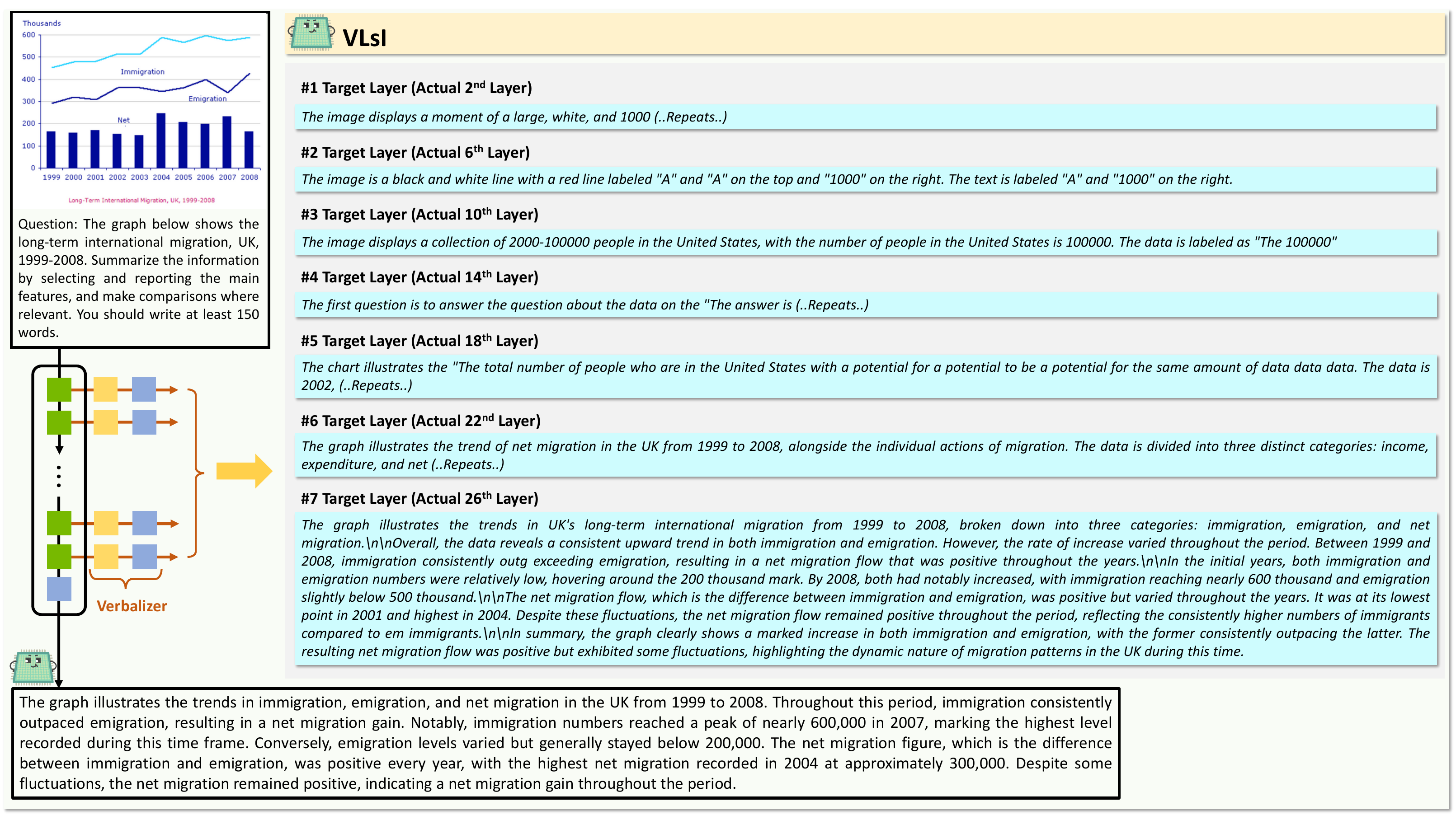}
\end{figure}

\begin{figure}[h!]
    \centering
    \includegraphics[width=\textwidth]{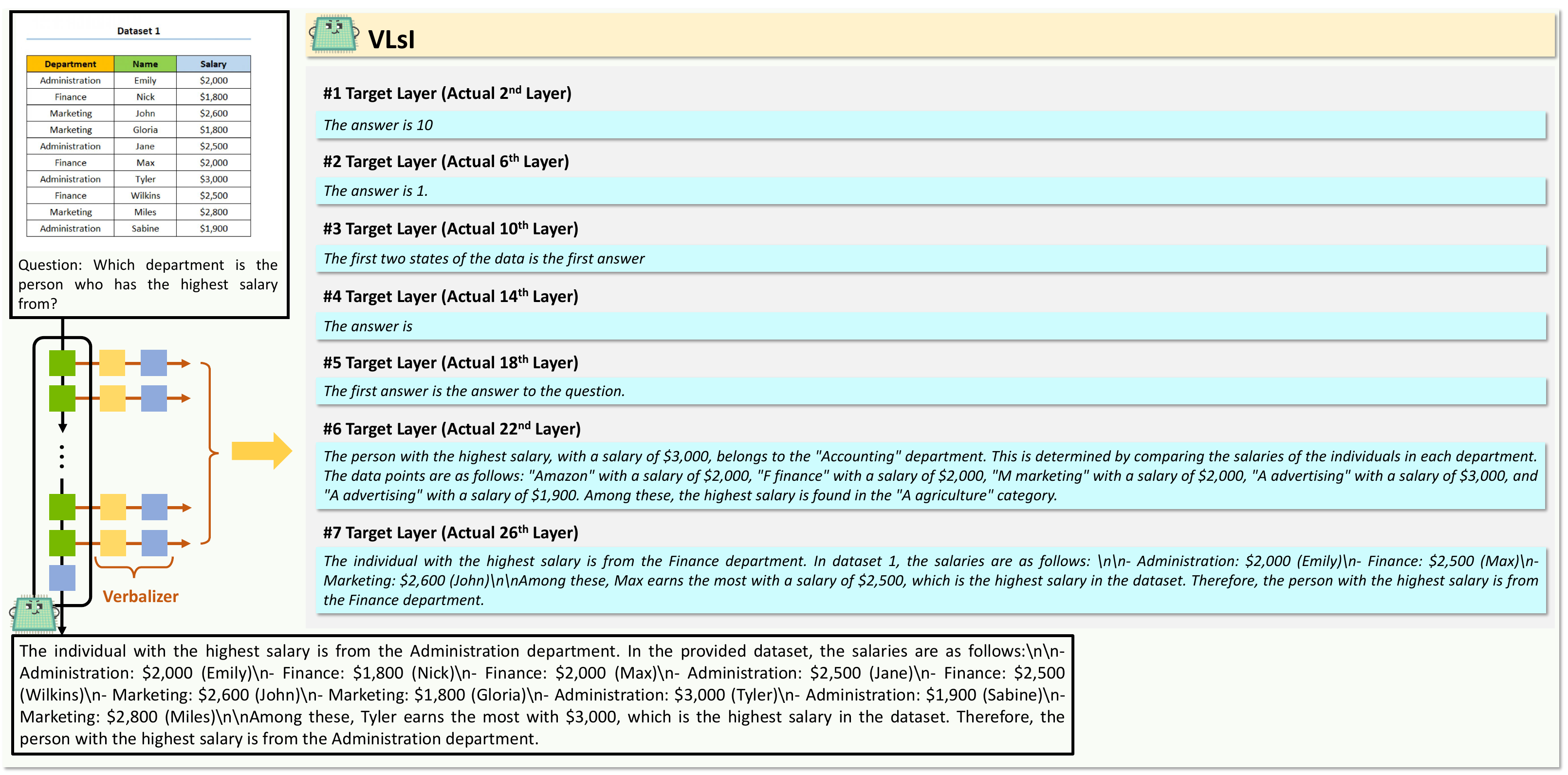}
\end{figure}

\begin{figure}[h!]
    \centering
    \includegraphics[width=\textwidth]{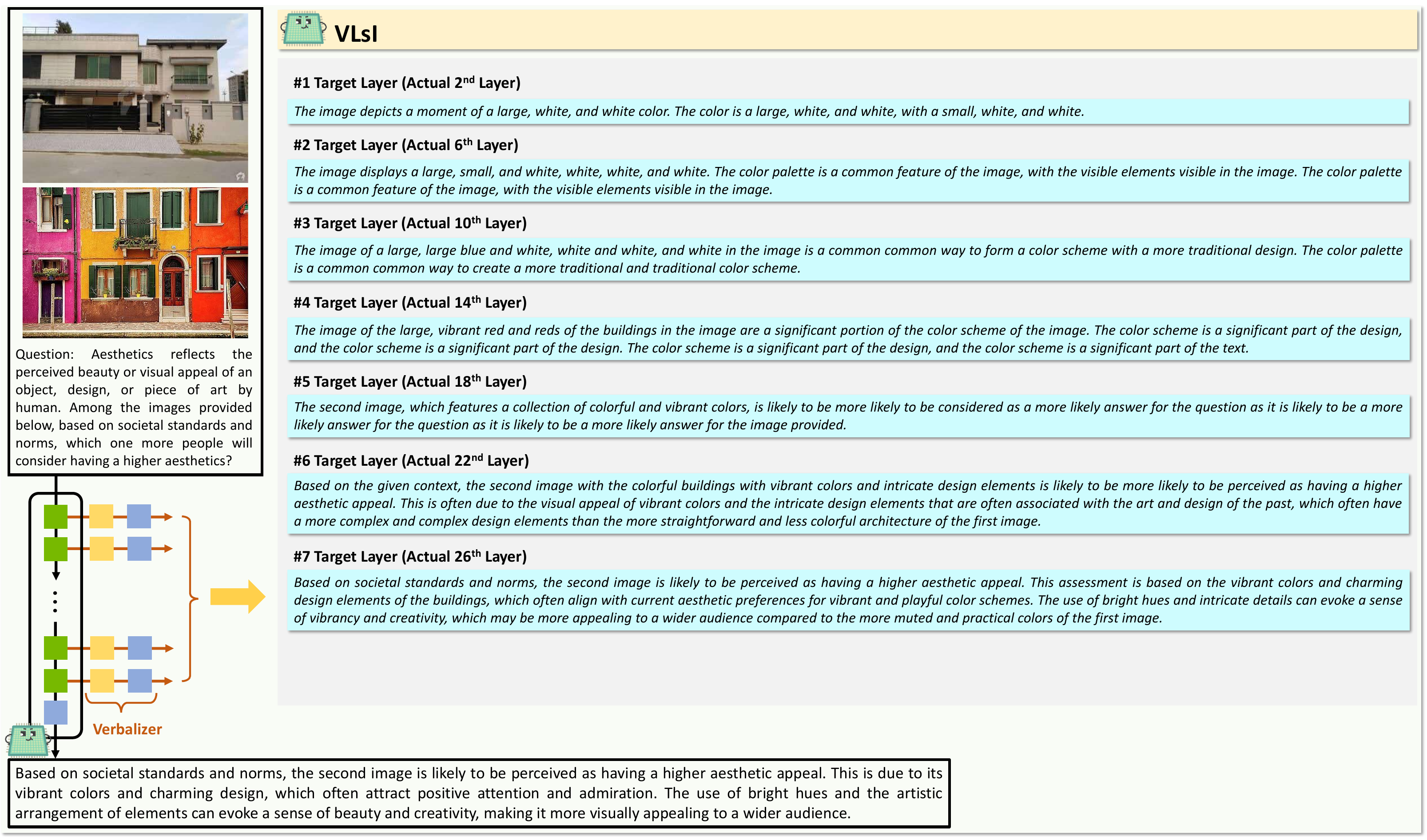}
\end{figure}

\clearpage
\section{Comprehensive Experimental Setup for Ablation Studies}
\label{app:D}

\paragraph{Training and Inference Configuration of LLaVA-OV~\cite{li2024llava}.} Different from Qwen2-VL~\cite{wang2024qwen2vl}-based \vlsi VLsI, we select intermediate target layers of LLaVA-OV-based \vlsi VLsI to $i_s$: 4$^{\text{th}}$, 12$^{\text{th}}$, 20$^{\text{th}}$ layers, and $i_t$: 10$^{\text{th}}$, 30$^{\text{th}}$, 50$^{\text{th}}$, and 70$^{\text{th}}$ layers. Although LLaVA-OV's language model is Qwen2~\cite{yang2024qwen2} that is equal with that of Qwen2-VL, the required number of image tokens in LLaVA-OV are 4 to 10 times more number than that of Qwen2-VL, depending on the height-to-width ratio of the images, given the same pixel count. To accommodate this increased computational demand on 8 NVIDIA A100 80GB GPUs, we reduce the number of intermediate target layers. For efficient training, we equally employ LoRA~\citep{hu2021lora} with rank and alpha parameters to 64, use the AdamW optimizer~\citep{loshchilov2018decoupled} with a cosine annealing schedule, adjusting the learning rate from $1e^{-4}$ to $1e^{-6}$ over each training step, and use gradient accumulation with 16 steps and gradient checkpointing~\citep{sohoni2019low}. The only different training configuration is batch sizes where two batches each are used for the 0.5B and 7B model sizes, and one batch is used for the 72B model sizes. It finally results in 256 (8×16×2) and 128 (8×16×1) batches, respectively. We conduct inference experiments under the equal setup used in Qwen2-VL, where we use a greedy search for text generation.

\begin{table}[h!]
\centering
\begin{tabular}{lcc}
\toprule
VLMs         & MM-Vet & MMMU \\
\midrule
LLaVA-OV-7B  & 57.5   & 48.8 \\
Qwen2-VL-7B  & \textbf{62.0}   & \textbf{54.1} \\
\midrule
LLaVA-OV-72B & 63.7   & 56.8 \\
Qwen2-VL-72B & \textbf{74.0}   & \textbf{64.5} \\
\bottomrule
\end{tabular}
\vspace{-5mm}
\end{table}

\paragraph{\cref{fig:4}} provides the challenging evaluation benchmarks' performances: MM-Vet~\cite{yu2023mm} and MMMU~\cite{yue2023mmmu}. Each cell represents the performances on their evaluation benchmarks, where the orange colored-values represent LLaVA-OV-based \vlsi VLsI's result and the purple ones represent Qwen2-VL-based result. This figure reveals consistent trends that using large- and small-backbone VLMs with more bigger model sizes enhances \vlsi VLsI's performances across all configurations. Besides, we can easily infer that the baseline performances of Qwen2-VL before buliding \vlsi VLsI are also higher than those of LLaVA-OV as shown in the above table. Furthermore, \cref{tab:4}(a) shows \vlsi VLsI's generalization ability to 0.5B and 7B model sizes in LLaVA-OV, but similarly insists that using better large-backbone VLMs provides benefits from performances.

\paragraph{\cref{tab:4}(b)} collectively illustrate the impact of incorporating the \textit{SFT step}. Table (b) highlights the significant performance gains achieved by applying additional SFT. Interestingly, the performances without additional SFT step are not satisfactory when doing greedy decoding, but once we utilize beam decoding ($N$=5), its performances are dramatically enhanced. This suggests that the interaction step implicitly alters and expands the probability space over the language head, while SFT aligns this expanded space to better fit instruction-following. Thus, significant improvements arise from the combination of interaction and SFT step. 

\paragraph{\cref{tab:4}(c)} It examines whether these improvements result from the fine-tuning effects of 2.9 million visual instruction tuning samples. The results demonstrate that \vlsi VLsI's performance becomes markedly superior to Qwen2-VL as SFT training progresses from 50\% to 100\%, confirming that additional SFT plays a much more critical role in driving performance enhancements than the contribution of the visual instruction tuning samples alone.

\begin{table}[h!]
\centering
\begin{tabular}{llcccc}
\toprule
IL-Ops                & LL-Ops  & vL-Head& MM-Vet & MMMU \\
\midrule
L2                    & \xmark  & \xmark & 63.3	  & 53.5 \\
L2                    & KLD     & \xmark & 64.6	  & 55.9 \\
L2                    & L2      & \xmark & 63.9	  & 54.4 \\
\midrule
\xmark                & L2      & -      & 65.2	  & 56.8 \\
\xmark                & KLD     & -      & 66.5	  & 57.5 \\
\midrule
KLD                   & KLD     & \cmark & 75.8   & 69.3 \\
\bottomrule
\end{tabular}
\vspace{-5mm}
\end{table}

\paragraph{\cref{tab:4}(d)} evaluates the impact of different operations applied to intermediate (IL-Ops) and last layers (LL-Ops) on \vlsi VLsI's performances. The results clearly demonstrate that KL divergence (KLD) operations applied to both intermediate and last layers yield the best performances, confirming the effectiveness of KLD over cross-entropy (CE) or L2 for intermediate layer alignment and final layer interaction. To further assess the effectiveness of verbalization in transferring the knowledge via distillation, we conduct an ablation study by removing the language head in verbalizer (vL-head). In this setup, the \textit{verbalization step} is skipped, and the \textit{interaction step} is first performed by aligning the hidden dimensions of the verb-FFNs in large- and small-backbone VLM. Here, the verb-FFN's hidden dimension in large-backbone VLM is kept but the hidden dimension in small-backbone VLM's verb-FFN is enlarged to match the large one. In addition, the verb-FFNs of large- and small-backbone VLM are interacted with only L2, which means that small-backbone VLM's verb-FFNs try to naively follow those of large-backbone VLM. After \textit{interaction step}, \textit{SFT step} is equally conducted. However, as shown in the above table, this naive approach results in significantly lower performance compared to the version equipped with the language head. Interestingly, using only last layer distillation by L2 and KLD provides more benefits than using intermediate distillation without vL-Head. These findings suggest that directly imitating outputs from the large-backbone VLM, without verbalization from vL-Head, introduces instability and can lead to suboptimal results, highlighting the critical role of the language head in achieving effective distillation.

\setlength{\textfloatsep}{0pt}
\begin{algorithm}
\caption*{Pseudo-Code for Random Index (Search Range \xmark)}
\begin{algorithmic}[1]
    \STATE \textbf{Input:} $t_s$, $t_l$
    \STATE \textbf{Initialize:} loss: 0, $i_{l}^{*}$: 0, $\epsilon$: 1e-6, scale: 2
    \FOR{$i_s$ in $0\leq i_s<t_s$}
        \STATE \textit{kld-list} = []
        \FOR{$i_{l}$ in $0\leq i_{l}< t_{l}$}
            \STATE \textit{kld-list}.append(\textit{compute-kld}($i_{s}$, $i_l$))
        \ENDFOR
        \STATE $r \gets$ Random-Select(\textit{kld-list})
        \STATE loss $\gets$ loss $+r$ 
    \ENDFOR
    \STATE \textbf{Return:} loss
\end{algorithmic}
\end{algorithm}

\setlength{\textfloatsep}{0pt}
\begin{algorithm}
\caption*{Pseudo-Code for Uniform Index (Search Range \xmark)}
\begin{algorithmic}[1]
    \STATE \textbf{Input:} $t_s$, $t_l$
    \STATE \textbf{Initialize:} loss: 0, $i_{l}^{*}$: 0, $\epsilon$: 1e-6, scale: 2
    \STATE $\text{layer-gap-ratio}\gets\text{floor}(\frac{t_l}{t_s})$
    \FOR{$i_s$ in $0\leq i_s<t_s$}
        \STATE $i_l\gets  i_s \times \text{layer-gap-ratio}$
        \STATE $u\gets$\textit{compute-kld}($i_{s}$, $i_l$)
        \STATE loss $\gets$ loss $+u$ 
    \ENDFOR
    \STATE \textbf{Return:} loss
\end{algorithmic}
\end{algorithm}

\paragraph{\cref{tab:4}(e)} highlights the effectiveness of various components in the matching strategy. Random Index, Uniform Index, and Bottom-1 or 3 Index, yield lower scores, underscoring the limitations of simpler selection mechanisms. Note that, the above and below algorithms represent the their detailed experimental setup. Multinomial sampling provides improvements and incorporating the Search Range further enhances performances (\textit{e.g.}, 83.5 on MMB and 69.8 on MM-Vet). Adding order preservation results in a significant leap, particularly on BLINK (59.2) and MM-Vet (75.2), demonstrating the importance of maintaining matched indices' sequence alignment during \textit{interaction step}. Finally, using all together with adaptive temperature achieves the best results across all benchmarks (e.g., 86.3 on MMB and 69.3 on MMMU), showcasing its ability to dynamically control the distribution. These exploration underscore the necessity of advanced sampling strategies and adaptive mechanisms for maximizing the efficiency of transferring the knowledge.

\clearpage

\setlength{\textfloatsep}{0pt}
\begin{algorithm}
\caption*{Pseudo-Code for Bottom-$k$ Index (Search Range \xmark)}
\begin{algorithmic}[1]
    \STATE \textbf{Input:} $t_s$, $t_l$
    \STATE \textbf{Initialize:} loss: 0, $i_{l}^{*}$: 0, $\epsilon$: 1e-6, scale: 2
    \FOR{$i_s$ in $0\leq i_s<t_s$}
        \STATE \textit{kld-list} = []
        \FOR{$i_{l}$ in $0\leq i_{l}< t_{l}$}
            \STATE \textit{kld-list}.append(\textit{compute-kld}($i_{s}$, $i_l$))
        \ENDFOR
        \STATE \textit{bottom-$k$-kld-list}$\gets$ Bottom-$k$(\textit{kld-list}) \COMMENT{length(\textit{bottom-$k$-kld-list}): $k$}
        \STATE $a \gets$ Average(\textit{bottom-$k$-kld-list})
        \STATE loss $\gets$ loss $+a$ 
    \ENDFOR
    \STATE \textbf{Return:} loss
\end{algorithmic}
\end{algorithm}

\paragraph{\cref{tab:4}(f)} highlights verb-FFN's efficiency among different verbalizer architectures in terms of performance and parameter count. While larger architectures such as Decoder×2 (3.3B) and FFN$\times2$ (2.9B) achieve strong performance across MM-Vet and MMMU, the much smaller verb-FFN architecture (269M) delivers comparable results. In contrast, simpler structures like MLP$\times2$ and MLP, while efficient in terms of parameter count, fall short in performance. These results emphasize the effectiveness of the verb-FFN architecture in a lightweight yet high-performing solution. Its efficiency becomes even more pronounced as the number of intermediate target layers increases and larger-backbone VLMs are utilized.

\clearpage
\section{VLsI's Text Generation Quality}
\label{app:E}
\begin{figure}[h!]
    \centering
    \includegraphics[width=0.8\textwidth]{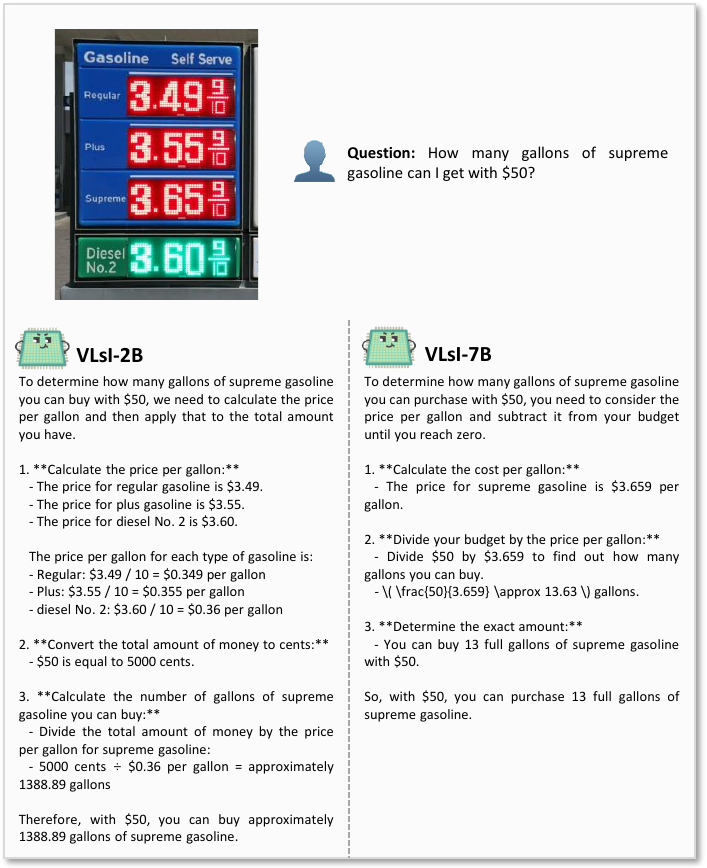}
\end{figure}

\begin{figure}[h!]
    \centering
    \includegraphics[width=0.8\textwidth]{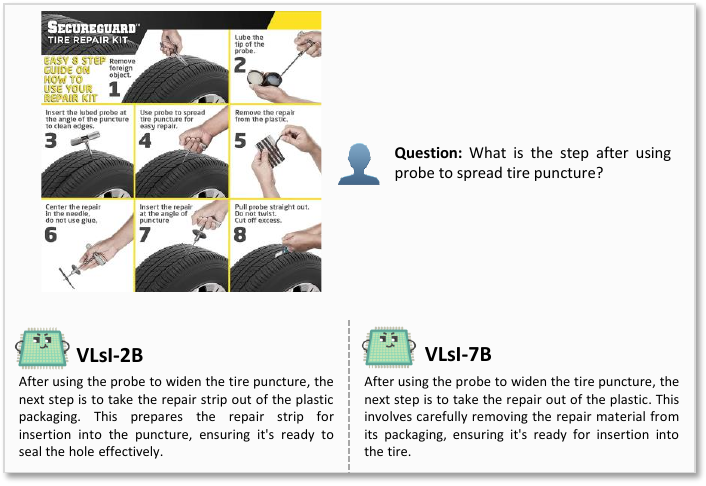}
\end{figure}

\begin{figure}[h!]
    \centering
    \includegraphics[width=0.8\textwidth]{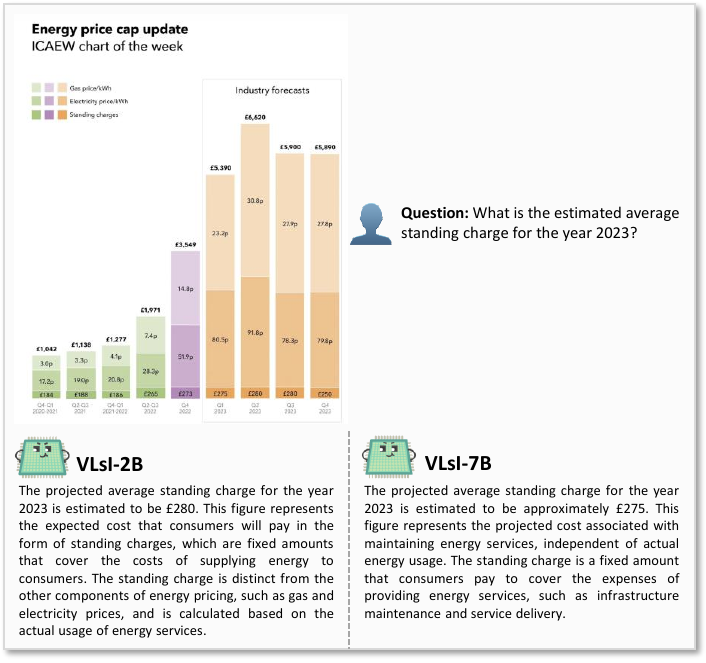}
\end{figure}

\begin{figure}[h!]
    \centering
    \includegraphics[width=0.8\textwidth]{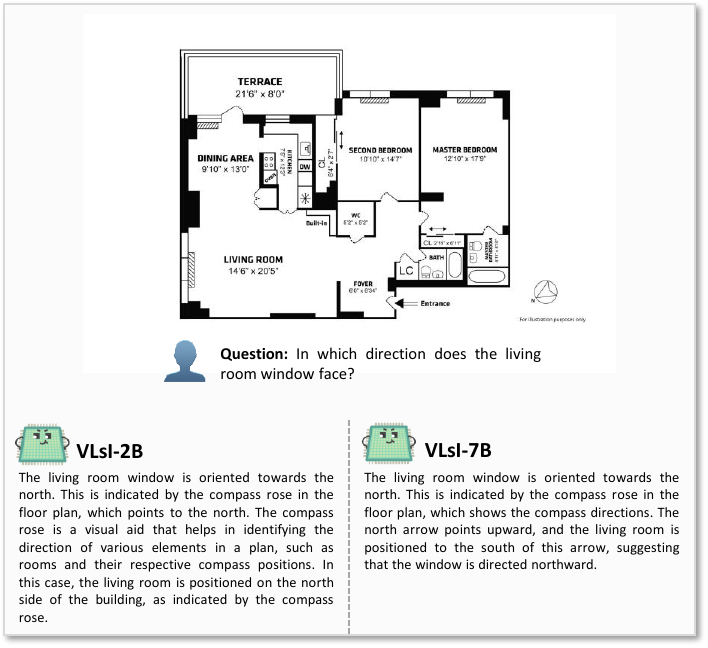}
\end{figure}

\begin{figure}[h!]
    \centering
    \includegraphics[width=0.8\textwidth]{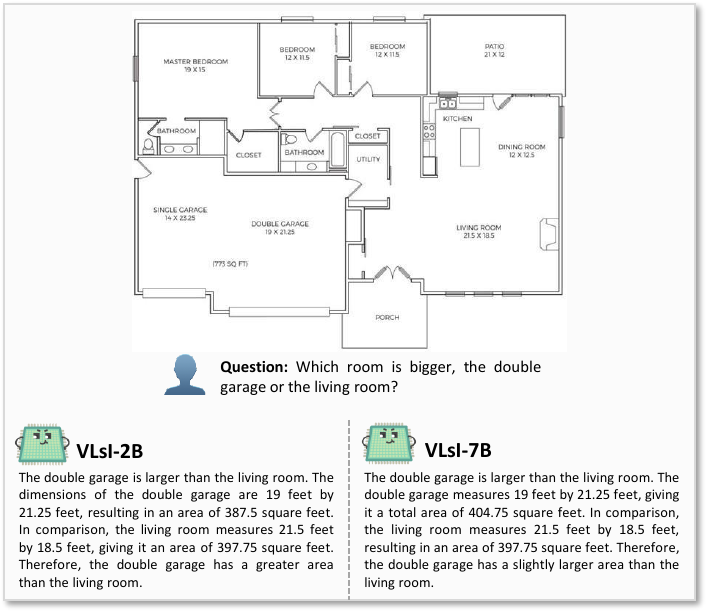}
\end{figure}

\begin{figure}[h!]
    \centering
    \includegraphics[width=0.8\textwidth]{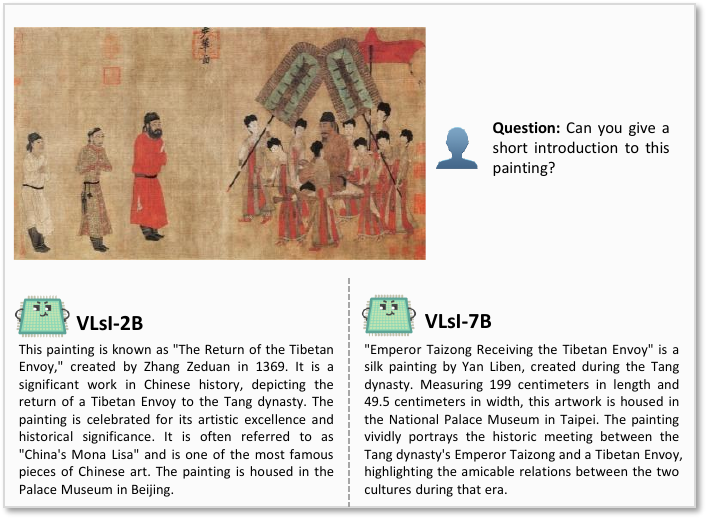}
\end{figure}

\begin{figure}[h!]
    \centering
    \includegraphics[width=0.8\textwidth]{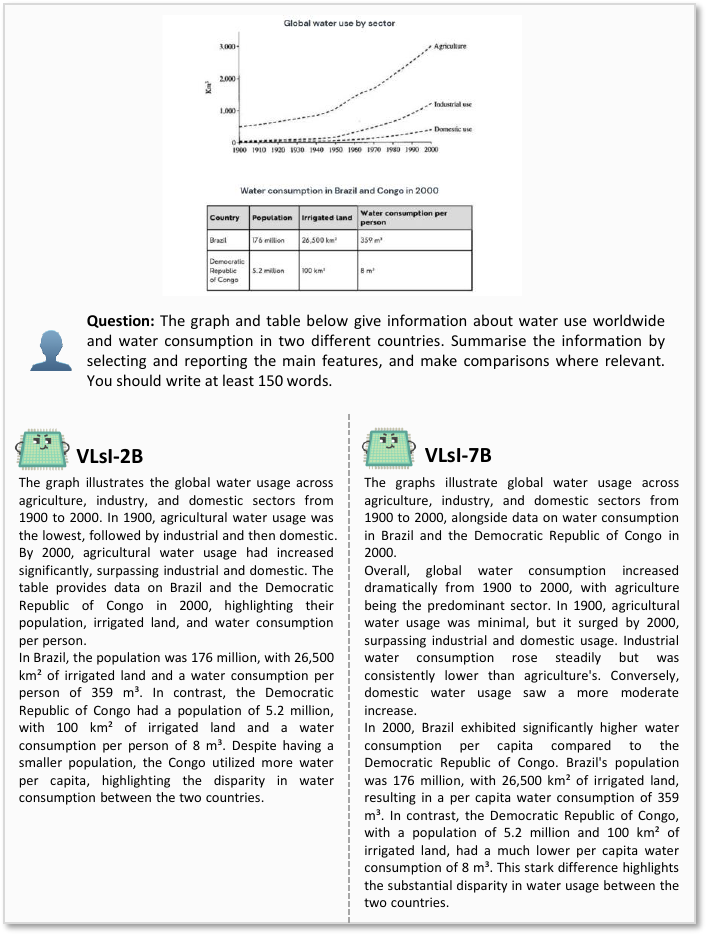}
\end{figure}

\begin{figure}[h!]
    \centering
    \includegraphics[width=0.8\textwidth]{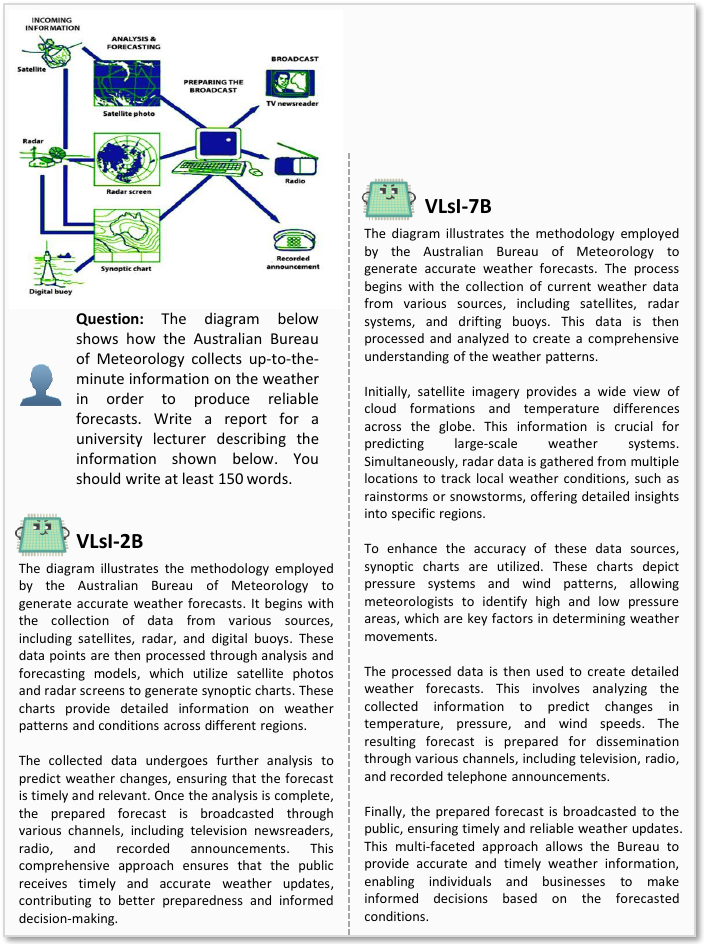}
\end{figure}

\begin{figure}[h!]
    \centering
    \includegraphics[width=0.9\textwidth]{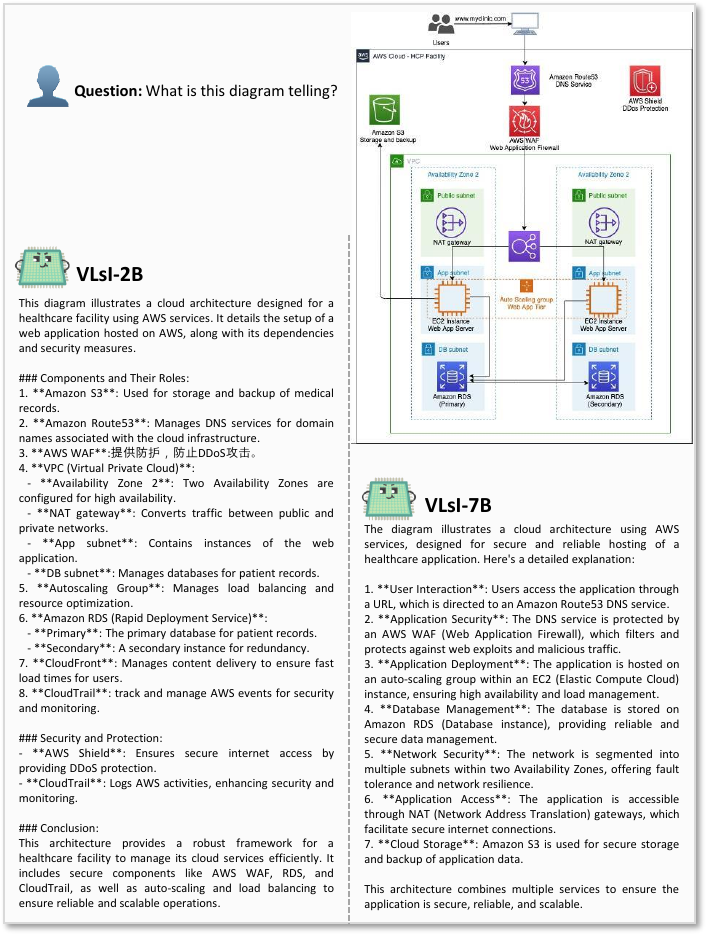}
\end{figure}

\begin{figure}[h!]
    \centering
    \includegraphics[width=0.8\textwidth]{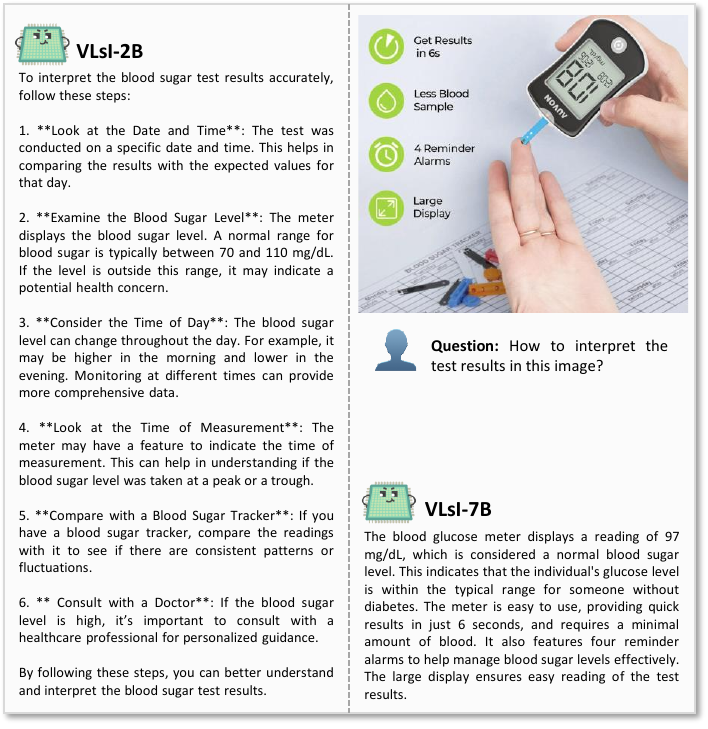}
\end{figure}

\begin{figure}[h!]
    \centering
    \includegraphics[width=0.8\textwidth]{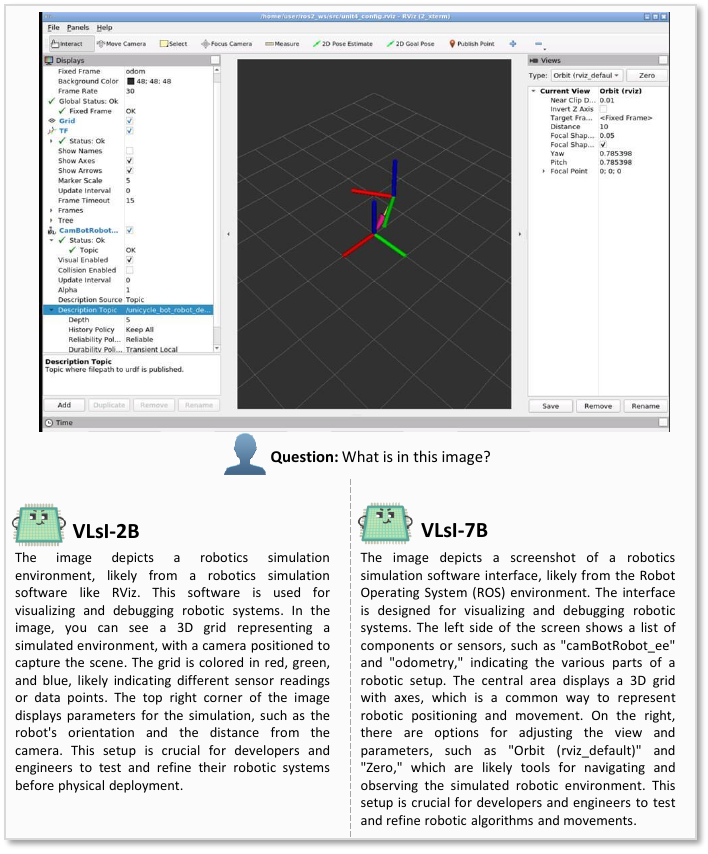}
\end{figure}

\begin{figure}[h!]
    \centering
    \includegraphics[width=0.8\textwidth]{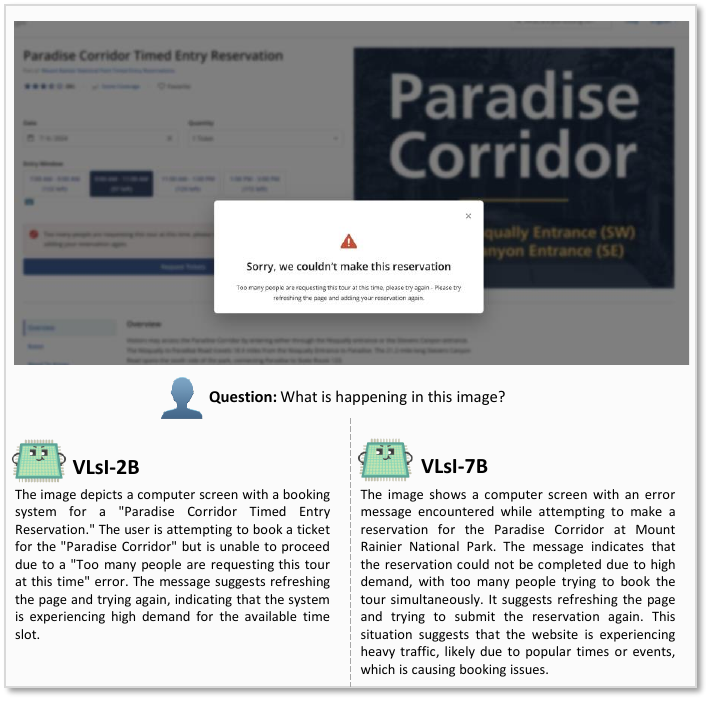}
\end{figure}

\end{document}